\documentclass[10pt,twocolumn,letterpaper]{article}

\usepackage{iccv}              

\usepackage{amsmath}
\usepackage{stmaryrd}
\usepackage{graphicx}
\usepackage{natbib}
\usepackage{multirow}
\usepackage{algorithm,algpseudocode}
\usepackage{wrapfig}
\usepackage{enumitem}

\definecolor{dark_blue}{HTML}{1a3e5c}

\definecolor{iccvblue}{rgb}{0.21,0.49,0.74}
\usepackage[pagebackref,breaklinks,colorlinks,allcolors=iccvblue]{hyperref}

\def\paperID{12232} 
\def\confName{ICCV}
\def\confYear{2025}

\title{LaCoOT: Layer Collapse through Optimal Transport}

\author{
Victor Quétu\textsuperscript{1} \quad
Zhu Liao\textsuperscript{1} \quad
Nour Hezbri\textsuperscript{2} \quad
Fabio Pizzati\textsuperscript{3} \quad
Enzo Tartaglione\textsuperscript{1} \\
\textsuperscript{1}LTCI, Télécom Paris, Institut Polytechnique de Paris, France\\
\textsuperscript{2}ENSAE, Institut Polytechnique de Paris, France\\
\textsuperscript{3}MBZUAI, UAE\\
{\tt\small victor.quetu@telecom-paris.fr}
}

\begin{document}
\maketitle
{\renewcommand{\thefootnote}{}%
\footnotetext{This paper has been accepted for publication at the IEEE/CVF International Conference on Computer Vision 2025 (ICCV25).}
}
\begin{abstract}
Although deep neural networks are well-known for their outstanding performance in tackling complex tasks, their hunger for computational resources remains a significant hurdle, posing energy-consumption issues and restricting their deployment on resource-constrained devices, preventing their widespread adoption.\\
In this paper, we present an optimal transport-based method to reduce the depth of over-parametrized deep neural networks, alleviating their computational burden. More specifically, we propose a new regularization strategy based on the Max-Sliced Wasserstein distance to minimize the distance between the intermediate feature distributions in the neural network. We show that minimizing this distance enables the complete removal of intermediate layers in the network, achieving better performance/depth trade-off compared to existing techniques.
We assess the effectiveness of our method on traditional image classification setups and extend it to generative image models. 
Our code is available at \url{https://github.com/VGCQ/LaCoOT}.
\end{abstract}
\section{Introduction}
\label{sec:intro}
Over the last few years, the field of deep learning has undergone a significant transformation with the advent of foundation models.
These are large-scale, pre-trained models capable of performing a wide range of tasks across different domains, including computer vision. As exemplars we can mention CLIP~\cite{radford2021learning} and ALIGN~\cite{jia2021scaling} for image classification, DiT~\cite{peebles2023scalable}, Stable Diffusion~\cite{rombach2022high} and DALL-E~\cite{ramesh2022hierarchical} for image generation, or SAM~\cite{kirillov2023segment} for semantic segmentation. 
The effectiveness of these foundation models is primarily driven by empirical patterns observed through scaling laws~\cite{hestness2017deep}: the improvements achieved by these models correlate with the exponential increase in computational requirements due to the growth of both their size and the number of training data~\cite{rosenfeld2021scaling,dehghani2023scaling}.

However, the progress enabled by these new models (consisting of billions of parameters) comes at the price of higher computational costs, consuming more energy, thus contributing to carbon emissions~\cite{MLSYS2022_462211f6}. For instance, training a generative model is comparable to driving a car for 10km, while generating 10k samples is estimated to be equivalent to driving 160 km~\cite{seyfarth2024latent}. Although training costs are expensive, the open-sourcing of these foundation models multiplies inference costs across multiple users and contributes to carbon emissions in a significant manner. The need to reduce the environmental impact of these models at inference by proposing computational reduction is therefore apparent.
\begin{figure}[t]
    \centering
    \includegraphics[width=\columnwidth]{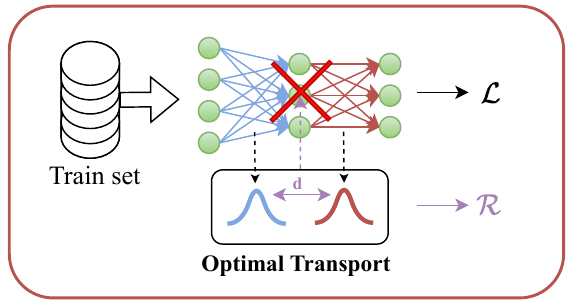}
    \caption{With LaCoOT, we finetune existing networks with an OT-inspired regularization $\mathcal{R}$ augmenting the loss $\mathcal{L}$, reducing intermediate feature distribution discrepancy. This enables the complete removal of layers.}
    \label{fig:teaser}
\end{figure}
Consequently, the rise of complexity-reduction approaches such as pruning~\cite{han2015learning}, quantization~\cite{gholami2022survey}, and knowledge distillation~\cite{hinton2015distilling} is motivated by the need for more efficient architectures to alleviate their resource demands. Reducing deep neural network (DNN) complexity is not an easy task: generalization and model complexity are inextricably related~\cite{hestness2017deep}, but since pre-trained models are often employed for downstream tasks, they tend to be over-parameterized. This gives us hope: in principle, it is possible to compress these models without any (or only little) performance degradation~\cite{tartaglione2022loss}. This observation is further supported by the \emph{collapse} phenomenon in neural networks, which has been observed at both the neuron~\cite{zhu2021a} and layer levels~\cite{gromov2024unreasonable}. On the one hand, individual neurons stop learning, leading to constant or trivial outputs, while on the other hand, entire layers fail to learn and become redundant or inactive. Consequently, these layers could, in principle, be removed.

Nevertheless, few methods are capable of removing entire layers from a neural network. 
Some of them have been designed to mitigate the depth of DNNs while maintaining performance, exploiting the deletion of several layers~\cite{liao2023can} or ad-hoc architectural search~\cite{baymurzina2022review}. However, these methods are computationally challenging since they either require retraining or rely on huge search spaces, leading to significant search costs. Moreover, the focus of previous works is often more on removing non-linearities, leaving the fusion of the remaining consecutive linear layers to further research, which shows that this is not straightforward in many common cases~\cite{unknown}.   

Driven by the motivation of reducing the depth of DNNs, we leverage optimal transport (OT)~\cite{villani2009optimal,peyre2019computational} to develop a framework allowing post-training the complete removal of layers from the architecture. Compared to existing OT-based frameworks incorporating it into neural architecture search~\cite{yang2023hotnas,nguyen2021optimal,NEURIPS2018_f33ba15e} or knowledge distillation pipelines~\cite{lohit2022model,chen2021wasserstein}, our strategy does not involve training more than one network but rather operates inside the model. In our case, we use OT to minimize the distributional changes between layers inside the same model, allowing us to strategically and efficiently remove layers (as showcased in Fig.~\ref{fig:teaser}).
Overall, our contributions can be summarized as follows.

\begin{itemize}[noitemsep,nolistsep]
    \item We propose a novel OT-based and block collapse inductive regularization (Sec.~\ref{sec:LaCoOT}), seamlessly integrated into the main training pipeline of neural networks. Our approach consists of minimizing a block-wise OT discrepancy measure, specifically the Max-Sliced Wasserstein distance, between the input and output features' probability distributions of the blocks of the network (Sec.~\ref{subsec:procedure}).
    \item We motivate our strategy (Sec.~\ref{sec:strategy} and Sec.~\ref{sec:properties}) by showing how it allows, post-training, the complete removal of several blocks from the architecture at once.
    \item Our proposed regularization strategy demonstrates its effectiveness in reducing the depth of over-parameterized DNNs with marginal performance loss with respect to competing state-of-the-art techniques (Sec.~\ref{results}).
\end{itemize}

\section{Related Works}
\label{sota}
\noindent \textbf{Neural network pruning.}
In the last decades, neural network pruning has risen as the one privileged approach to compress deep neural networks: complimentary to other popular approaches like quantization, it leads to heavy parameter reduction through the proper cut of groups of parameters (or filters in convolutional architectures) that are less important for the specific downstream task under exam. Its effectiveness is empirically certified by several works~\cite{blalock2020state,cheng2024survey,he2023structured} and justified by the known overparametrization of such models~\cite{liebenwein2021lost}. 
Among these, we historically distinguish between \emph{unstructured} pruning approaches that eliminate parameters without considering the neural network's structure~\cite{han2015learning,tartaglione2022loss} and \emph{structured pruning}, where entire channels, neurons, or filters are removed~\cite{he2023structured,tartaglione2021serene}.

Unstructured pruning methods are grouped into two main categories based on the nature of the importance score used to prune weights: \emph{gradient}-based methods rank the parameters according to the gradient magnitude~\cite{lee2018snip,tartaglione2022loss} (or higher-order derivatives), while \emph{magnitude}-based ones~\cite{han2015learning,louizos2018learning,h.2018to} use the weights' magnitude as a significance score to prune them. In a famous study, \cite{blalock2020state} compared the effectiveness of these two approaches, concluding that magnitude-based techniques are often more accurate than gradient-based ones while offering a better trade-off between complexity and competitiveness. Following up on this work, \cite{Gale_Magnitude} even showcased that simple magnitude pruning methods can achieve results that are comparable to more complex ones, establishing a solid comparison baseline. 
Although some studies suggest that unstructured pruning may actually harbor structured effects~\cite{liao2024nepenthe}, in general, they provide few practical benefits when deploying the neural network on generic computing resources~\cite{bragagnolo2021role}.

Unlike unstructured pruning, structured pruning brings immediate advantages for both memory and computation, despite resulting in lower overall sparsity~\cite{bragagnolo2021role}. 
Despite this, when employing recent computing resources, removing entire filters on recent computing resources has only a marginal effect on the improved latency, given the availability of resources in parallel. The real bottleneck in parallel computation resides in the computational \emph{critical path}\footnote{We refer to the critical path as the longest path, in terms of time or computational cost, through the computational graph that must be executed sequentially during inference, thus determining the model’s minimum achievable latency.}~\cite{mehmeti2023nonlinear}, which can be mitigated by reducing the model's depth. Although some existing approaches, like knowledge distillation to shallow student models~\cite{hinton2015distilling} already tackle this issue, maintaining the performance cannot be guaranteed, as the optimal architecture of the target model is not known a priori, which may lead to significant performance loss. 

\noindent \textbf{Neural network depth reduction.}
Several recent works have proposed approaches to reduce the depth of DNNs. The most common practice is to \emph{remove non-linearities} between layers: this (in principle) enables two successive layers to be merged together. Among these approaches, Layer Folding~\cite{dror2021layer} is one of the earlier attempts: it evaluates whether non-linear activations can be discarded, replacing ReLUs with PReLU (having a trainable slope for the negative part). More recently, Entropy-Guided Pruning (EGP)~\cite{liao2023can} proposes to reduce the depth of DNNs by prioritizing the pruning of connections in layers with little use of the non-linearity (estimated through an entropic measure). On the same trend, NEPENTHE~\cite{liao2024nepenthe} improved EGP's entropy estimator and introduced a budget for the number of parameters to prune. 
Taking a more global approach, EASIER~\cite{quetu2024simpler} was designed to determine the effect of removing a non-linearity, considering the introduced error at the output of the model, estimated through a validation set. 

Although in principle effective in reducing the critical path length of DNNs, these methods do not provide significant gains in all cases, due to some impossibilities in merging consecutive linear layers. One example is the problem associated with ResNet-type architectures: if we have padding in the second convolutional layer, then there is no analytical solution for merging the two consecutive layers~\cite{unknown}. Moreover, if an activation is removed where there is a residual connection, no fusion can be applied. A major architectural change is necessary to observe a real impact in practice.

Unlike those works, our method does not rely on linearizing activations and merging two consecutive layers. Instead, LaCoOT minimizes the Max-Sliced Wasserstein distance between the input and output features’ distributions of the blocks of the network. Post-training, layers having the lowest Max-Sliced Wasserstein distance are removed. Developed to be model-agnostic, we compare our method with these works and show its effectiveness in Sec.~\ref{sec:results}. 
\section{Preliminaries}
\label{sec:preliminaries}

\subsection{Background on Optimal Transport}
\label{sec:background}
In this subsection, we present a succinct overview of OT and the Wasserstein distance for discrete distributions~\cite{peyre2019computational}.

Given two metric spaces $\mathcal{X}$  and $\mathcal{Y}$ and a cost function $c$ defined over $\mathcal{X}\times \mathcal{Y} $, the goal of the OT problem is to determine the most efficient manner to transport mass from one distribution, defined over $\mathcal{X}$ to another supported over $\mathcal{Y}$, where the transportation cost is dictated by the chosen function $c$.  

For $\mathcal{X}= \mathcal{Y}= \mathbb{R}^d$, we consider two discrete probability measures and we recall the \emph{Monge Formulation of the OT problem}:
\begin{equation}
    OT(\mu, \nu , c) = \min_{T} \sum_{i}{\alpha_{i}c[\boldsymbol{x}_i, T(\boldsymbol{x}_i)] } \text{,}
\end{equation}
where $\mu $ and $ \nu$ defined as:  
\begin{align}
    \mu= \sum_{i=1}^{N}\alpha_{i}\delta_{\boldsymbol{x_{i}}},\qquad
    \nu= \sum_{i=1}^{M}{\beta_{i}\delta_{\boldsymbol{y}_{i}}},
\end{align}
where $\delta_{x}$ refers to the Dirac (unit mass) distribution at point $\boldsymbol{x}$. The weights $\alpha $ and $\beta $  reside in the probability simplex $\{ a \in \mathbb{R} | \sum { a_i} = 1\}$, and $T$ is defined as $~{ T: \{ \boldsymbol{x}_1, \dots , \boldsymbol{x}_N \} \rightarrow \{ \boldsymbol{y}_1, \dots , \boldsymbol{y}_M \} } $ and verifying: 
\begin{equation}
    \beta_j= \sum_{i: T(\boldsymbol{x}_i)=\boldsymbol{y}_j} \alpha_i, \quad \forall j \in \llbracket M \rrbracket,
\end{equation}
or more compactly $ ~{T_{\sharp}\mu=\nu }$. \\
In the following, we will consider only the case of uniform weights and the same support size, taking $~{M=N \text{ and } ~{\alpha_i= \beta_j= \frac{1}{N}}}$. We also take as the cost function $~{c(\boldsymbol{x}, \boldsymbol{y}) = \| \boldsymbol{x}-\boldsymbol{y} \|^{p}_{p}}$ for $~{\boldsymbol{x} \in \mathcal{X}}, ~{\boldsymbol{y} \in \mathcal{Y},} ~{p\in\mathbb{R}_{>0}}$. In this case, OT establishes a measure of distance between the probability distributions. Such a distance, known as the \textbf{\( \textbf{p}\)-Wasserstein distance}, is in general defined as $~{\mathcal{W}_{p} = OT(\mu, \nu, c)^{\frac{1}{p}}}$. 
When we have the dimension of the ground space being \( d = 1 \), the p-Wasserstein distance takes on a closed form, given by:
\begin{equation}
    \mathcal{W}_{p} = \left( \frac{1}{N} \sum_{i=1}^{N} \left| \boldsymbol{x}_{i} - \boldsymbol{y}_{i} \right|^p \right)^{\frac{1}{p}},
\end{equation}
where we assume $x_1<\dots< x_N$ and $ y_1<\dots< y_N$ such that $\boldsymbol{x}_{i}\mapsto \boldsymbol{y}_i, \forall i $.\\
Given the closed-form expression in one dimension, sliced variants of the p-Wasserstein distance have been introduced. These variants transform sample assignment and distance calculation by sorting the one-dimensional projection of the samples. This process yields a sufficient approximation of the high-dimensional p-Wasserstein distance, which is immune to the curse of dimensionality~\citep{curse_dim}. Specifically, our focus lies on the \textbf{p-Max-Sliced Wasserstein distance}, introduced in~\cite{Deshpande2019}, and defined as follows: 
\begin{equation}
   \max\!~\tilde{W}_p(\mu, \nu) = \underset{\theta \in \mathcal{U}(\mathbb{S}^{d-1})} {\max} \mathcal{W}_p(\theta_{\sharp}\mu, \theta_{\sharp}\nu),
\end{equation}
where $ \theta_{\sharp} $ stands for the pushforwards of the projection $ X: \mathbb{R}^d \mapsto \left \langle \theta, X \right \rangle $ , $\left \langle \cdot,\cdot\right \rangle $ for the dot product operator and $\mathcal{U}(\mathbb{S}^{d-1})$ for the uniform distribution on the unit hyper-sphere of dimension $d-1$.

Essentially, the Max-Sliced Wasserstein distance represents a version of the sliced Wasserstein distance where we select the optimal direction to project the probability measures, \ie , the direction along which the projected distance is maximized, also possessing valid metric properties~\citep{Nadjahi2020StatisticalAT,nadjahi2021fast,bonneel:hal-00881872}. 
In our work, we will consider the Max-Sliced Wasserstein distance, for its previously discussed convenience, specifically computed for $p=2$, to quantify the distance between intermediate probability distributions between blocks inside a neural network model, as it will be presented in the next section.

\subsection{Learning Framework}
\label{sec:notation}
In this subsection, we introduce our learning framework. 
Let us define \( \mathcal{T} = T_{K} \circ \dots \circ T_{1} \) as the DNN we wish to train on the dataset \( \mathcal{D} \), where each \( T_{k} \) is an elementary module (which can be defined as single or multiple layers). Given a loss function \( \mathcal{L} \) we aim at minimizing, the objective entails minimizing the problem:
\begin{equation}
    (\mathcal{T}, F) \in \arg\min_{\mathcal{T},F}\sum_{i=1}^{|\mathcal{D}|} \mathcal{L}\left[(F \circ \mathcal{T})(\boldsymbol{x}_{1,i}), y_i\right],
\end{equation}
where $F$ is a classifier layer, $\boldsymbol{x}_{1,i}$ the $i$-th input sample for the DNN, and $y_i$ the associated ground-truth label and $|\mathcal{D}|$ the number of samples in the dataset. For each module \( T_k \), we define the \emph{input probability distribution} \( \mu_k \) as:
\[
\mu_k := \frac{1}{|\mathcal{D}|} \delta_{\boldsymbol{y}_{k-1, \mathcal{D}}} = \frac{1}{|\mathcal{D}|} \delta_{\boldsymbol{x}_{k, \mathcal{D}}} = \frac{1}{|\mathcal{D}|} \sum_{i=1}^{|\mathcal{D}|} \delta_{\boldsymbol{x}_{k, i}},
\]
which is also the output of the preceding module \( T_{k-1} \).
Similarly, the \emph{output probability distribution} \( \nu_k \) is defined as:
\[
\nu_k := \frac{1}{|\mathcal{D}|} \delta_{\boldsymbol{y}_{k, \mathcal{D}}} = \frac{1}{|\mathcal{D}|} \sum_{i=1}^{|\mathcal{D}|} \delta_{\boldsymbol{y}_{k, i}}.
\]
According to our notation, $\nu_k\equiv \mu_{k+1}$, given that $\boldsymbol{y}_{k,i} = \boldsymbol{x}_{k+1,i}\forall i,k$. 
Furthermore, we assume that \(\boldsymbol{x}_{k,i} \) and \(\boldsymbol{y}_{k,i} \) possess identical dimensions, and consequently \( \mu_{k} \) and \( \nu_{k} \) to live in the same dimensional space, allowing for the computation of the distance between them. In the following, our goal would be to control these distances during training to allow for the isolation and removal of certain blocks post-training, with almost no loss of performance.

\section{LaCoOT}
\label{sec:LaCoOT}
In this section, we detail our method, LaCoOT, for reducing neural network depth using optimal transport (Sec.~\ref{sec:strategy}): we propose a regularization strategy based on the Max-Sliced Wasserstein distance to minimize the distance between intermediate feature distributions in the neural network. Fig.~\ref{fig:teaser} provides a general overview of our method. We also present some insights into this strategy and how it enables the removal of intermediate layers in the network after training, with, in principle, minimal performance degradation (Sec.~\ref{sec:properties}).

\subsection{Proposed Regularization}
\label{sec:strategy} 
Our objective is to reduce the depth of the neural network. To achieve such a goal, we will incorporate a penalization of the distance between two consecutive blocks during training. This is to assist the DNN in learning a target input-output function $\mathcal{T}$ while following the shortest path. This yields, after training, to identifying blocks that can be removed from the architecture without impacting the performance. Specifically, these blocks introduce marginal statistical modification on their corresponding input features.
We include this constraint in the learning translates into minimizing, besides the loss $\mathcal{L}$, in the form of a regularizer:
\begin{equation}
    \mathcal{R} = \frac{1}{K} \sum_{k=1}^{K} \max\!~\tilde{W}_2(\hat{\mu}_{k}, \hat{\nu}_{k}) \text{,}
    \label{eq:regu}
\end{equation}
where the probability distributions \( \hat{\mu}_{k} \) and  \( \hat{\nu}_{k} \)  are the empirical counterparts of the previously defined distributions \( \mu_{k} \) and \( \nu_{k} \). They are constructed over uniformly-weighted samples of a $N$-sized minibatch, and so defined by \( \hat{\mu}_{k} = \frac{1}{N} \sum_{i=1}^{N}{\delta_{\boldsymbol{x}_{k,i}}} \), where \( \boldsymbol{x}_{k,i} \) is taken as the flattened input vector of the block corresponding to the \( i \)-th element of the minibatch.

Post-training, if the distance \( \widehat{\mathcal{R}}_k:=\max\!~\tilde{W}_2 (\hat{\mu}_{k}, \hat{\nu}_{k}) \) falls below a fixed threshold \( \varepsilon \), the corresponding block \( T_{k} \) can be pruned from the architecture. Namely, this threshold is related to a tolerated performance drop budget~$\delta$.

\subsection{Overview on the Procedure}
\label{subsec:procedure}
\begin{algorithm}[t]
    \caption{Our proposed method LaCoOT.}
    \label{alg}
    \begin{algorithmic}[1]
        \Function{{\texttt{LaCoOT}}($\boldsymbol{w}^{\text{INIT}}$, $\mathcal{D}$, $\lambda$, $\delta$)}{}
        \State $\boldsymbol{w}\gets $Train($\boldsymbol{w}^{\text{INIT}}$, $\mathcal{D}_{\text{train}}$, $\lambda$)~\label{line:train_dense}
        \State dense\_acc $\gets $Evaluate($\boldsymbol{w}$, $\mathcal{D}_{\text{val}}$)\label{line:evaluate_dense}
        \State current\_acc $\gets$ dense\_acc
        \While{(dense\_acc - current\_acc) $>$ $\delta$} 
        \State $\widehat{\mathcal{R}} = [\widehat{\mathcal{R}}_1,\widehat{\mathcal{R}}_2, ..., \widehat{\mathcal{R}}_K]$ \label{line:evaluate_wasserstein} 
        \State $l$ $\gets$ argmin($\widehat{\mathcal{R}}$)\label{line:argmin}
        \State $T_l$ = Identity() \label{line:linearize}  
        \State current\_acc $\gets$ Evaluate($\boldsymbol{w}$, $\mathcal{D}_{\text{val}}$)\label{line:re-evaluate}
        \EndWhile
        \State \textbf{return} $\boldsymbol{w}$
        \EndFunction
    \end{algorithmic}
\end{algorithm}

Depicted in Alg.~\ref{alg}, we present here LaCoOT to remove the layers having the lowest Max-Sliced Wasserstein distances. 
Indeed, the layer having the lowest Max-Sliced Wasserstein distance is likely to have a function close to the identity function.
Therefore, this layer can be linearized, as keeping it is unnecessary, as illustrated in Tab.~\ref{tab:indicator} in Supp. Mat. 
Aiming at this, we first train the neural network, represented by its weights at initialization $\boldsymbol{w}^{\text{INIT}}$ on the training set $\mathcal{D}_{\text{train}}$ with our regularization set by $\lambda$ (line~\ref{line:train_dense}) and evaluate it on the validation set $\mathcal{D}_{\text{val}}$ (line~\ref{line:evaluate_dense}).
We then calculate the Max-Sliced Wasserstein distance $\widehat{\mathcal{R}}_k$ for each considered layer $k$ for all the $K$ considered layers, collected in the vector $\mathcal{R}$ (line~\ref{line:evaluate_wasserstein}), following Eq.~\ref{eq:regu}.
We then find the layer having the lowest Max-Sliced Wasserstein distance, represented by its index $l$ (line~\ref{line:argmin}) and replace it with the Identity (line~\ref{line:linearize}).
In the following steps, this layer is, obviously, no longer taken into consideration. 
The performance of the model is re-evaluated on the validation set $\mathcal{D}_{\text{val}}$ (line~\ref{line:re-evaluate}). 
Once the performance on the validation set drops below the threshold $\delta$, the final model is obtained.

\subsection{Properties of the proposed regularization}
\label{sec:properties}

\textbf{The regularization acts like a soft 1-Lipschitz constraint.} 
By looking closer into $\mu_k$ and $\nu_k$ (the input and output distributions of the $k$-th block), the central limit theorem suggests that $\mu_k$ can be regarded asymptotically as a Gaussian distribution with mean $\boldsymbol{m}_k$ and covariance $\Sigma_k$. Then, by employing the delta method, $\nu_k$ can be approximated asymptotically as a Gaussian distribution with mean $T_k(\boldsymbol{m}_k)$ and covariance $J_k^T\Sigma_k J_k$, where $J_k$ represents the Jacobian matrix of the block transformation $T_k$.

The constraint  $\mu_k=\nu_k$ can be interpreted as an orthogonality constraint on the Jacobian of the block transformation, indicating that our regularization imposes a similar effect as enforcing orthogonality on the Jacobian. This implies a block-wise soft Lipschitz constraint on the neural network by preserving gradient norms that drive the network to be 1-Lipschitz. This type of constraint was investigated in the literature~\citep{Li2019PreventingGA, Anil2018SortingOL,thune2022pay}, and it has been particularly shown in~\cite{thune2022pay} that a 1-Lipschitz constraint does not limit the expressiveness, \ie the capacity and learning flexibility of a neural network, for classification tasks. Instead, this regularization should offer a different stance on the trade-off between generalization and accuracy. This also coincides with the results in~\cite{leastact} that highlight the generalization-enhancing effect of such a regularization. During training, provided that a proper weight on the regularization (through some hyperparameter $\lambda$) is tuned, the neural network's expressive power should, in principle, remain intact, while adhering to the least action principle, thereby preventing arbitrary amplification of small differences and big distributional changes. 

\noindent \textbf{Stationary point analysis.} 
\label{parag: gradient_analysis}
The whole optimization problem can be expressed as:
\begin{equation}
   \mathcal{J}= \mathcal{L}+\lambda\mathcal{R} \Rightarrow  \frac{\partial \mathcal{J} }{\partial w_{k_0}}= 
     \frac{\partial \mathcal{L} }{\partial w_{k_0}}+ \lambda  \frac{\partial \mathcal{R} }{\partial w_{k_0}} \text{,} \quad k_0 \in \llbracket K\rrbracket,
     \label{eq:objective}
\end{equation}
where $\lambda$ is a positive hyperparameter. We then characterize the stationary point as: 
\begin{equation}
     \frac{\partial \mathcal{L} }{\partial w_{k_0}}+ \lambda  \frac{\partial \mathcal{R} }{\partial w_{k_0}}=0
     \Rightarrow \lambda=-\frac{\partial \mathcal{L} }{\partial w_{k_0}} \cdot \frac{1}{\frac{\partial \mathcal{R} }{\partial w_{k_0}}}.
\end{equation}
From this equation, since $ \lambda \geq 0 $, we clearly observe that the loss and the regularizer are antagonists. Hence, while browsing the parameter space to minimize the loss, unrestricted DNNs are rather biased towards increasing intermediate distributional changes in the path they take. These changes, which can be evaluated when taking the inter-block distances in the vanilla setting, might be irrelevant: the DNN can converge to another local minimum in the loss landscape, with similar performance but without undergoing too many distributional changes.

In the learning process, we recall that the primary objective is to traverse the gap between the input distribution and the target output distribution. A crucial threshold is thus reached when the network's output converges to the ground truth. Namely, this inherent distance between the input and ground truth distribution defines a tight lower bound for the regularization value, corresponding to the minimal distributional changing capacity that still has to be maintained in the network to have a good performance and not to underfit. This is guaranteed by applying the Triangle inequality: 
\begin{align}
\max\!~\tilde{W}_2(\mu_1, \nu_{GT})\leq \sum_{k=1}^K& \max\!~\tilde{W}_2(\mu_k, \nu_k) \nonumber\\&+ \max\!~\tilde{W}_2(\nu_K, \nu_{GT}) \text{,}
 \end{align}
where $\nu_{GT}$ represents the ground truth label distribution.
\section{Experiments}
\label{sec:results}

In this section, we empirically evaluate the effectiveness of our proposed approach across multiple architectures and datasets for traditional image classification setups and extend it to image generation. 

\subsection{Experimental setup}
\label{exp.setup}
\begin{figure*}[t]
    \centering
    \includegraphics[width=\textwidth]{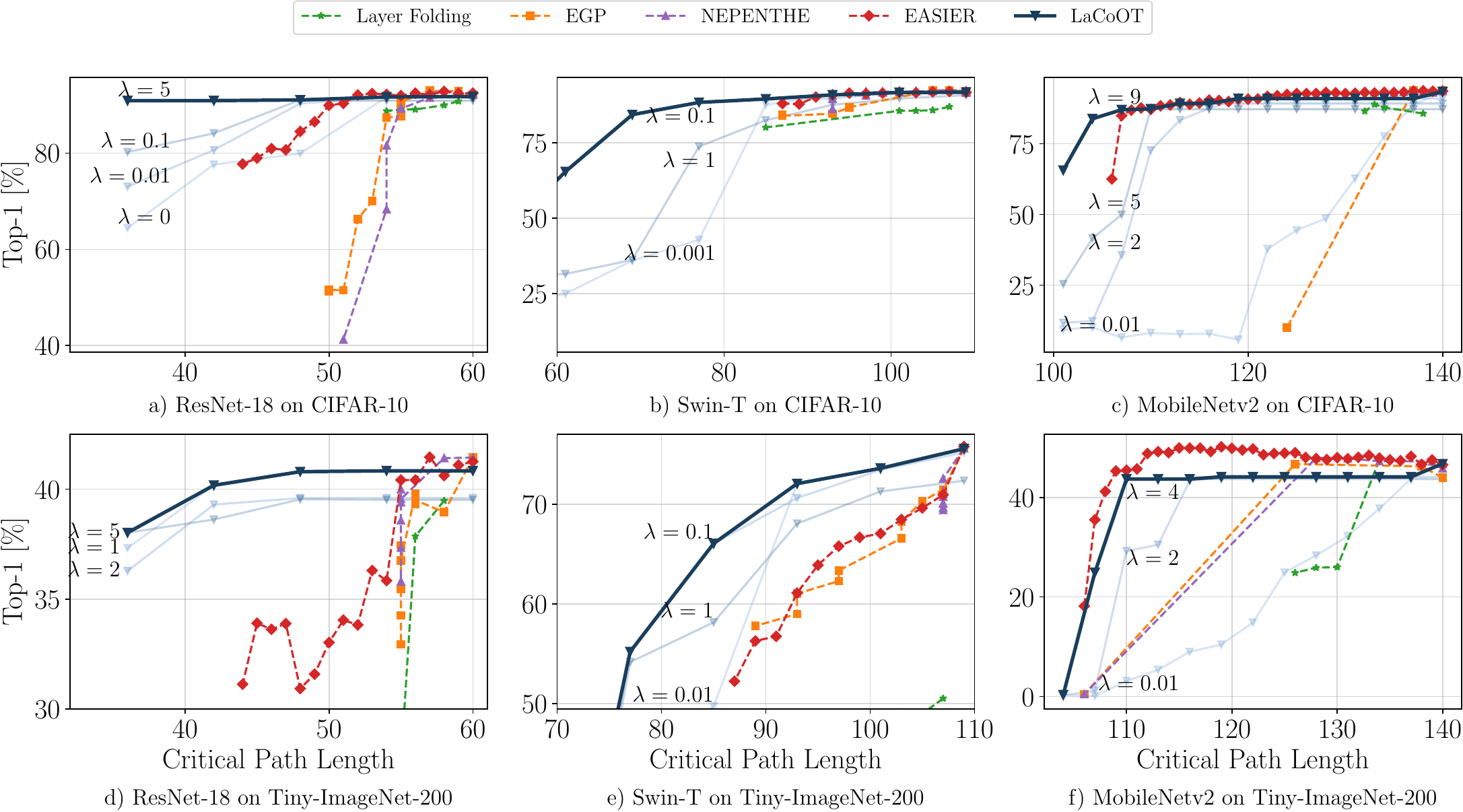}
    \caption{Test performance (Top-1 [\%]) in function of the Critical Path Length for ResNet-18~(a,d), Swin-T~(b,e) and MobileNetv2 (c,f) trained on CIFAR-10~(a,b,c) and Tiny-ImageNet-200~(d,e,f). For each dataset/architecture, we showcase the results achieved by LaCoOT for different values of $\lambda$, forming in \textbf{\textcolor{dark_blue}{dark blue}} the pareto frontier of our technique. \emph{Top left corner is the best.}}
    \label{fig:main_res}
\end{figure*}
\textbf{Networks and Datasets.} On image classification setups, we test LaCoOT on three widely used models, ResNet-18, MobileNet-V2, and Swin-T trained on seven different datasets: CIFAR-10~\cite{krizhevsky2009learning}, Tiny-ImageNet~\cite{le2015tiny}, PACS and VLCS from DomainBed~\cite{gulrajani2020search}, as well as Flowers-102~\cite{Nilsback08}, DTD~\cite{cimpoi14describing}, and Aircraft~\cite{maji13fine-grained}. To showcase its applicability to larger models on diverse tasks, we employ DiT-XL/2~\cite{peebles2023scalable}, finetuned on ImageNet~\cite{ImageNet} for image generation.
For all our experiments, we use the implementation of the Max Sliced Wasserstein distance available in the POT toolbox~\citep{pot}.
LaCoOT is only applied in subsequent blocks having the same dimensionality: this results in a subset of 4, 12, 12, and 28 blocks considered removable for ResNet-18, Swin-T, MobileNetV2, and DiT-XL/2, respectively.

\noindent\textbf{Baselines.}
We compare our method with leading depth-reducing methods: Layer Folding~\cite{dror2021layer}, EGP~\cite{liao2023can}, NEPENTHE~\cite{liao2024nepenthe}, and EASIER~\cite{quetu2024simpler}.
The hyperparameters, augmentation techniques, and learning policies are presented in Supp. Mat., mainly following~\cite{quetu2024dsd2} and~\cite{quetu2024simpler}.

\subsection{Results}
\label{results}
\subsubsection{Image Classification}
Fig.~\ref{fig:main_res} displays the test performance (top-1) as a function of the critical path length for CIFAR-10 and Tiny-ImageNet-200. The results achieved on other setups can be found in Sec.~\ref{appendix:more_results} in the Supp. Mat. Moreover, Fig.~\ref{fig:CPL} in Supp. Mat. illustrates the relationship between Critical Path Length (CPL) and practical resource consumption: as the CPL decreases, both the inference time and the number of MACs (multiply-accumulate operation) decrease, indicating improved computational efficiency and faster inference speeds.

\noindent \textbf{Critical Path Length.} First, we can observe in Fig.~\ref{fig:main_res} that the baseline methods showcase longer critical path lengths with respect to our method. Indeed, as discussed in Sec.~\ref{sota}, most methods have been focusing on removing non-linearities from the networks, leaving the fusion of subsequent layers as future work.  However, while for Swin-T the fusion of two linear layers in the MLP block is straightforward, it is not the case when ResNet is employed: we recall that there is no analytical solution for merging two consecutive convolutional layers when padding is employed in the second one~\cite{unknown}.
Hence, even if multiple non-linearities are removed from the network with these baselines, the critical path length only slowly decreases. Unlike these methods, LaCoOT focuses directly on full blocks divided by skip connections, hence quickly lowering the critical path length.
\begin{figure}[t]
    \centering
    \includegraphics[width=0.95\linewidth]{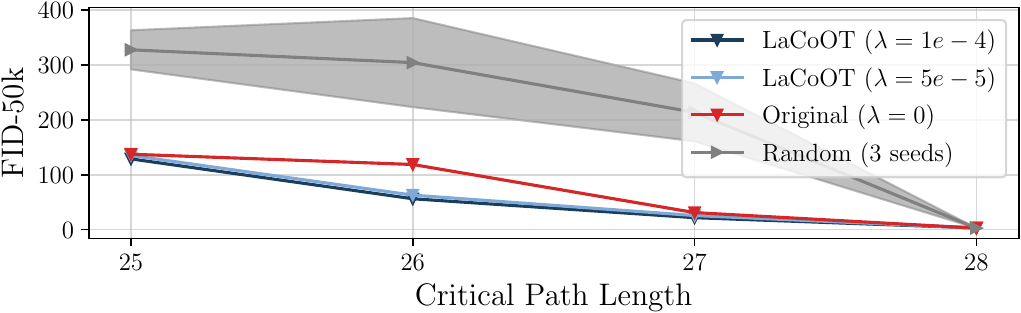}
    \begin{minipage}[b]{0.05\columnwidth}
        \centering
        \rotatebox{90}{~\textbf{\scriptsize{$\lambda=1e\!\mathpunct{-}\!4$}}~~~~~~~\textbf{\scriptsize{$\lambda=0$}}~~~~~~~~\scriptsize{Pre-trained}}
    \end{minipage}
    \begin{minipage}[b]{0.88\columnwidth}
        \centering
        \begin{subfigure}{\textwidth}
            \centering
        \includegraphics[width=\columnwidth]{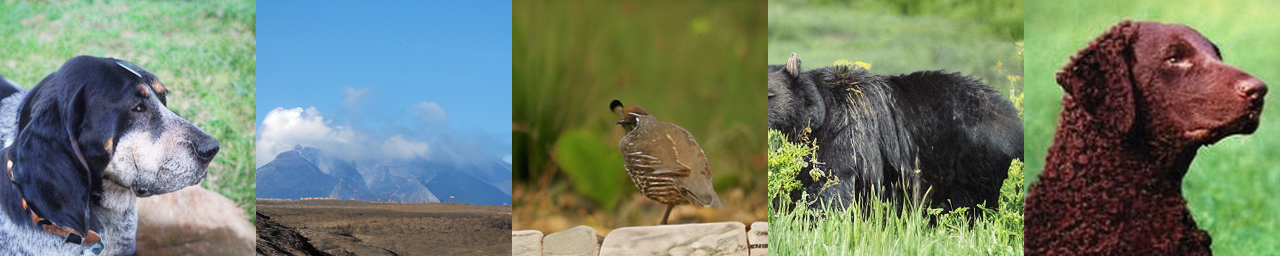}
        \end{subfigure}
        \hrule height 1.2pt
        \begin{subfigure}{\columnwidth}
            \centering
            \includegraphics[width=\columnwidth]{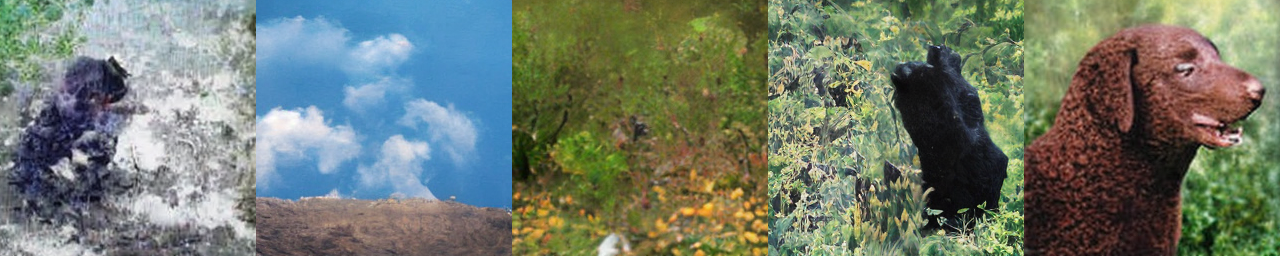}
        \end{subfigure}
        \begin{subfigure}{\textwidth}
            \centering
        \includegraphics[width=\columnwidth]{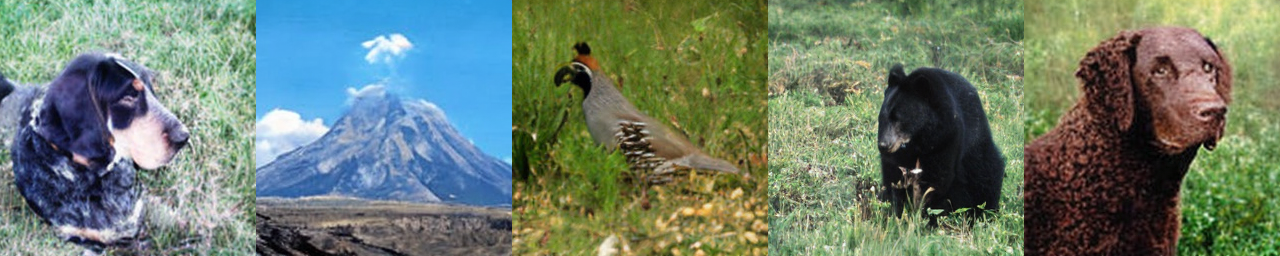}        
        \end{subfigure}
    \end{minipage}
    \begin{minipage}[b]{0.05\columnwidth}
        \centering
        \rotatebox{90}{~~\textbf{\tiny{$FID{=}56,2$}}~~~\textbf{\tiny{$FID{=}118,6$}}~~~~\tiny{$FID{=}2,9$}}
    \end{minipage}
    \caption{FID-50k as a function of the critical path length achieved by a DiT-XL/2 finetuned on ImageNet. It consistently achieves lower FID when finetuned with LaCoOT, even halving the FID when two DiT blocks are removed. The generated content is also better preserved (images for critical path length 26).}
    \label{fig:DiT-FID}
\end{figure}

\noindent \textbf{LaCoOT effectiveness.} 
First, in most setups, we can observe LaCoOT's effectiveness. Indeed, for short critical path length, LaCoOT is the method performing overall the best, achieving a new Pareto Frontier when $\lambda$ is increasing. For instance, for ResNet-18 trained on CIFAR-10 (Fig.~\ref{fig:main_res}a), our method reduces the critical path length even further, whereas previous methods could not. Besides, for Swin-T on Tiny-ImageNet-200 (Fig.~\ref{fig:main_res}e), LaCoOT sometimes outperforms current methods by 10\% for the same critical path length. Additionally, looking at longer critical path lengths, we can observe that when $\lambda$ is decreasing, LaCoOT achieves comparable results to other baselines. 

We also report an issue faced with EGP. Indeed, by forcing a layer to have zero entropy, this method could prune it entirely, hence preventing the signal from passing through this layer, and thus causing the algorithm to completely fail. This is what is observed with the MobileNetv2 architecture on CIFAR-10, Flowers-102, DTD, or on Aircraft: from the first iteration, EGP prunes the last single layer before the classifier head entirely, leading to its complete removal, which completely cuts the information flow in the network, since there is no skip or residual connection at this stage.

Furthermore, on parameter-efficient architectures (such as MobileNetv2), we can observe that EASIER performs the best (Fig.~\ref{fig:main_res}f), while our method achieves comparable results (Fig.~\ref{fig:main_res}c).
However, the advised reader will be able to put these results into perspective with the aim of EASIER, which focuses solely on removing non-linearities (as mentioned in Sec.~\ref{sota}). Moreover, the iterative nature of EASIER results in very few benefits in practice, as shown in Sec.~\ref{subsec:pratical}.
For instance, to achieve a path length of 105 on MobileNetv2, EASIER needs to carry out 34 trainings, whereas our method requires only one. 

\noindent \textbf{Comparison with the original model.}
Although in certain setups, such as ResNet-18 on CIFAR-10, LaCoOT effectively reduces model size while maintaining the original model’s performance, it often results in some performance degradation compared to the original model. This is likely because the model is not re-trained after layer removal. In contrast, traditional compression schemes typically involve re-training the model after dropping some parameters to recover performance. We show in Tab.~\ref{tab:NAS_like} in Supp. Mat. that the model can recover performance with a healing phase.  

\subsubsection{Image Generation}

\begin{table}[t]
    \centering
    \resizebox{\columnwidth}{!}{%
    \begin{tabular}{c|c|c|c|c}
    \toprule
        \textbf{Approach} & \textbf{top-1 [\%]} & \textbf{MACs [M]} & \textbf{Inference time [ms]} & \textbf{Time} \\
    \toprule 
        Original & 91.77 & 140.19 & 7.90 $\pm$ 0.43 & 30' \\
    \midrule 
        Layer Folding & 88.76 &  147.53 & 9.89 $\pm$ 1.11 & 160' \\
        EGP & 90.64 & 140.19 & 7.62 $\pm$ 0.20 & 376' \\
        NEPENTHE & 89.26 & 140.19 & 7.71 $\pm$ 0.40 & 288' \\
        EASIER & 90.35 & 140.19 & 7.07 $\pm$ 0.18 & 533' \\
    \midrule
        LaCoOT & \bf 90.99 & \bf 64.69 & \bf 4.78 $\pm$ 0.34 & \bf 40' \\ 
    \bottomrule
    \end{tabular}
    }
    \caption{Test performance (top-1), MACs, inference time on a NVIDIA A4500 and training time for ResNet-18 trained on CIFAR-10. Original refers to the trained model without layer deletion. The best results between Layer Folding, EGP, NEPENTHE, EASIER, and LaCoOT are in \textbf{bold}.}
    \label{tab:C10-R18_ablation}
\end{table}

Unlike other baselines, which cannot scale due to their iterative nature, we show the possible extension of our method to foundation models by finetuning a pre-trained DiT-XL/2 on ImageNet with our method LaCoOT for 5k training steps. For 50k generated samples of size 256$\times$256 with a classifier-free guidance scale of 1.5, Fig.~\ref{fig:DiT-FID} displays the FID-50k score depending on the critical path length, as well as examples of generated samples from the pre-trained model, and from models with two DiT blocks removed.

While no difference is observed when looking at the FID-50k score for the original model (critical path length of 28), the effect of our technique is visible when DiT blocks are removed: LaCoOT consistently achieves a lower FID-50k compared to the pre-trained model. For instance, when two DiT blocks are removed, we observe that the FID-50k score is twice as low.
This is reflected in the quality of the generated images: the dog, the volcano, the bird, and the bear are not really visible, while the last dog contains artifacts.
Indeed, the removal of blocks completely destroys generated images in the absence of the regularization, while the generated content is better preserved with its use. Therefore, as it required only a few fine-tuning steps, our approach LaCoOT can be suitable for foundation models. 

\subsection{Practical Benefits}
\label{subsec:pratical}
In this subsection, we showcase the practical benefits of our approach in terms of inference time as well as the efficiency of LaCoOT.\\
Tab.~\ref{tab:C10-R18_ablation} shows the test performance, MACs, and inference time on an NVIDIA A4500, as well as the training time for a ResNet-18 trained on CIFAR-10 for all the considered approaches. The same analysis for Swin-T and MobileNetv2 is conducted in Sec.~\ref{appendix:practical} in the Supp. Mat. 
As anticipated in Sec.~\ref{sota}, we observe that baseline methods do not reduce MACs in practice, as they simply rely on non-linearities removal without providing insights on how to merge consecutive layers. Unlike its competitors, LaCoOT produces a model whose inference has been reduced by 40\%. Moreover, looking at the training time, LaCoOT is the most efficient. 
In Tab.~\ref{tab:C10-MNv2_ablation} in the Supp. Mat., the same analysis is conducted for MobileNetv2 on CIFAR-10, corresponding to Fig.~\ref{fig:main_res}c. While Layer Folding, EGP, and NEPENTHE showcase performance drop at high critical path length, we achieve comparable performance as EASIER for the same critical path length in 20$\times$ less time, with real practical benefits since the inference time is reduced. 

\subsection{Ablation Study}
\label{ablation}

\begin{figure}[t]
    \centering
    \includegraphics[width=\columnwidth]{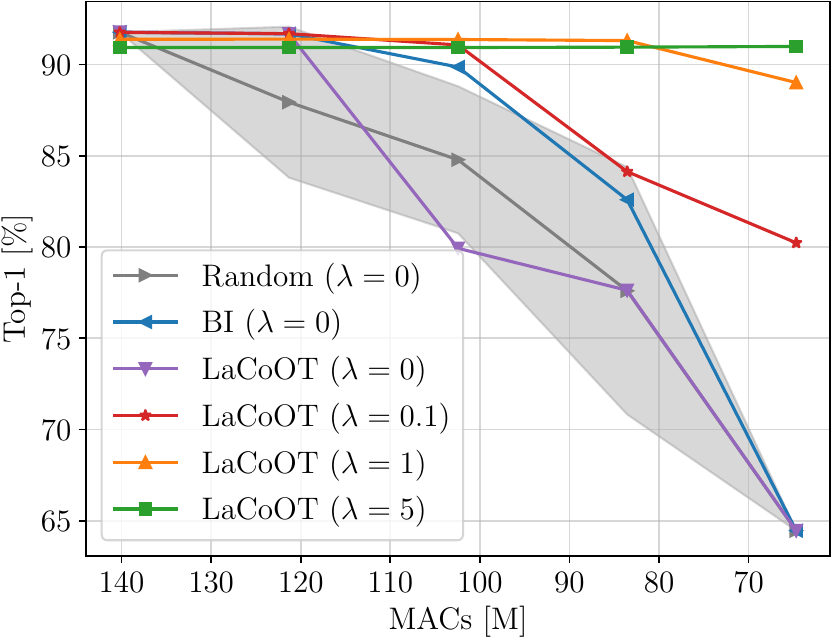}
    \caption{Comparison of LaCoOT, BI (theoretical best), and Random for ResNet-18 on CIFAR-10. LaCoOT ($\lambda = 5$) halves the MACs with minimal performance loss. Higher $\lambda$ values further reduce MACs while maintaining performance.}
    \label{fig:c10_r18_ablation}
\end{figure}

In this subsection, we study the impact of $\lambda$, which balances the strength of our regularizer. Moreover, to validate the effectiveness of our proposed importance metric for layer removal, we compare it with two alternative methods for selecting which layers to remove.
Fig.~\ref{fig:c10_r18_ablation} shows this comparison on a ResNet-18 on CIFAR-10.\\
\noindent \textbf{Impact of $\lambda$.} First, in the absence of regularization during training (\ie, with $\lambda=0$), we can observe that the Max-Sliced Wasserstein distance is not a faithful indicator of block importance, since it can be surpassed by random block removal.
Considering our previous theoretical analysis in Sec.~\ref{parag: gradient_analysis}, this observation is largely expected, since without our regularization, the blocks operate changes on the intermediate features' distribution, which is unnecessary. Indeed, when our regularization is incorporated into the training process, the unnecessary distributional changes are minimized, and our metric becomes a reliable basis for ranking the importance of the model blocks. Looking at Fig.~\ref{fig:c10_r18_ablation}, the higher $\lambda$, the more blocks can be removed without harming performance. Indeed, we can observe that LaCoOT ($\lambda = 5$) halves the number of MACs with almost no performance drop to the dense model. From Eq.~\ref{eq:objective}, selecting $\lambda$ depends on the user's goal, since there is a trade-off between performance and complexity. In general, the higher the lambda, the more layers can be removed, but \textit{potentially} the lower the performance.

\noindent \textbf{Choice of the importance score.} 
LaCoOT is here compared with two alternative methods for selecting which layers to remove. 
In the first scenario that we refer to as ``block influence'' (BI), we remove layers depending on their impact on the performance : we remove one layer at a time by selecting the one impacting the performance the least. 
In a second scenario, referred to as ``Random'', we remove one layer at a time by selecting it randomly. Based on randomness, error bars have been calculated on 10 seeds.
From Fig.~\ref{fig:c10_r18_ablation}, we can observe that LaCoOT achieves a better performance/compression trade-off compared to the two other approaches Block Influence and Random, which validates the choice of our importance score. 

\subsection{Limitations and Future Work}
\label{limitations}

While effective in alleviating the computational burden of DNNs, LaCoOT also has some limitations, highlighting opportunities for future improvements and research, as discussed below.

\noindent \textbf{Performance degradation.} Compressing existing parameter-efficient architectures is particularly challenging, a common issue in the field of model compression. Indeed, LaCoOT struggles to reduce the depth of an already underfitted architecture without compromising performance, as seen with MobileNetv2 on Tiny-ImageNet-200. This highlights the challenges of further compressing already efficient architectures and the need to carefully manage the trade-off between model depth and performance. However, LaCoOT effectively reduces the depth of over-fitted DNNs, especially given that only one training is required to achieve compression.

\noindent \textbf{Extension of LaCoOT with the Gromov-Wasserstein distance.}
LaCoOT was primarily designed to operate on layers where the Max-Sliced Wasserstein distance can be computed directly. While this distance requires matching dimensions between distributions, our experiments in Sec.~\ref{appendix:mismatch} in Supp. Mat. show that LaCoOT remains effective even for layers with mismatched dimensions, such as convolutional layers that modify the number of filters or feature sizes.
However, we believe that a more principled treatment of such cases—potentially leveraging the Gromov-Wasserstein distance~\cite{NEURIPS2019_a9cc6694}, which allows for comparing distributions whose supports do not necessarily lie in the same metric space—remains an avenue for future research.
\vspace{-0.cm}
\section{Conclusion}
\label{cl}

In this work, we have proposed LaCoOT, a new optimal transport-based regularization strategy. Specifically, we use the Max-Sliced Wasserstein distance to minimize the distances between the intermediate feature distributions in the neural network. 
This regularization enables, post-training, the complete removal of layers from the architecture with a minor impact on performance.
Experiments conducted on three widely used architectures across seven image classification datasets have demonstrated LaCoOT's capability and effectiveness in reducing the number of layers in the neural network. Unlike other approaches that rely on an iterative scheme, we have shown that extending LaCoOT to foundation models is possible since our approach requires only a few fine-tuning steps. 
Concerned about the increasing environmental impact of AI, we hope this work will inspire future optimization techniques and new approaches for network compression. 

\section*{Acknowledgements}
This work was supported by the French National Research Agency (ANR) in the framework of the JCJC project “BANERA” under Grant ANR-24-CE23-4369, from the European Union’s Horizon Europe Research and Innovation Programme under grant agreement No. 101120237 (ELIAS), and by the Hi! PARIS Center on Data Analytics and Artificial Intelligence. Zhu Liao acknowledges financial support from the China Scholarship Council (CSC).

{
    \small
    \bibliographystyle{ieeenat_fullname}
    \bibliography{main}

\begin{thebibliography}{65}
\providecommand{\natexlab}[1]{#1}
\providecommand{\url}[1]{\texttt{#1}}
\expandafter\ifx\csname urlstyle\endcsname\relax
  \providecommand{\doi}[1]{doi: #1}\else
  \providecommand{\doi}{doi: \begingroup \urlstyle{rm}\Url}\fi

\bibitem[Anil et~al.(2018)Anil, Lucas, and Grosse]{Anil2018SortingOL}
Cem Anil, James Lucas, and Roger~Baker Grosse.
\newblock Sorting out lipschitz function approximation.
\newblock In \emph{ICML}, 2018.

\bibitem[Baymurzina et~al.(2022)Baymurzina, Golikov, and Burtsev]{baymurzina2022review}
Dilyara Baymurzina, Eugene Golikov, and Mikhail Burtsev.
\newblock A review of neural architecture search.
\newblock \emph{Neurocomputing}, 474:\penalty0 82--93, 2022.

\bibitem[B{\'e}thune et~al.(2022)B{\'e}thune, Boissin, Serrurier, Mamalet, Friedrich, and Sanz]{thune2022pay}
Louis B{\'e}thune, Thibaut Boissin, Mathieu Serrurier, Franck Mamalet, Corentin Friedrich, and Alberto~Gonzalez Sanz.
\newblock Pay attention to your loss : understanding misconceptions about lipschitz neural networks.
\newblock In \emph{NeurIPS}, 2022.

\bibitem[Blalock et~al.(2020)Blalock, Gonzalez~Ortiz, Frankle, and Guttag]{blalock2020state}
Davis Blalock, Jose~Javier Gonzalez~Ortiz, Jonathan Frankle, and John Guttag.
\newblock What is the state of neural network pruning?
\newblock \emph{MLSys}, 2020.

\bibitem[Bonneel et~al.(2015)Bonneel, Rabin, Peyr{\'e}, and Pfister]{bonneel:hal-00881872}
Nicolas Bonneel, Julien Rabin, Gabriel Peyr{\'e}, and Hanspeter Pfister.
\newblock {Sliced and Radon Wasserstein Barycenters of Measures}.
\newblock \emph{{Journal of Mathematical Imaging and Vision}}, 2015.

\bibitem[Bragagnolo et~al.(2021)Bragagnolo, Tartaglione, Fiandrotti, and Grangetto]{bragagnolo2021role}
Andrea Bragagnolo, Enzo Tartaglione, Attilio Fiandrotti, and Marco Grangetto.
\newblock On the role of structured pruning for neural network compression.
\newblock In \emph{ICIP}, 2021.

\bibitem[Chen et~al.(2021)Chen, Wang, Gan, Liu, Henao, and Carin]{chen2021wasserstein}
Liqun Chen, Dong Wang, Zhe Gan, Jingjing Liu, Ricardo Henao, and Lawrence Carin.
\newblock Wasserstein contrastive representation distillation.
\newblock In \emph{CVPR}, 2021.

\bibitem[Cheng et~al.(2024)Cheng, Zhang, and Shi]{cheng2024survey}
Hongrong Cheng, Miao Zhang, and Javen~Qinfeng Shi.
\newblock A survey on deep neural network pruning: Taxonomy, comparison, analysis, and recommendations.
\newblock \emph{IEEE PAMI}, 2024.

\bibitem[Cimpoi et~al.(2014)Cimpoi, Maji, Kokkinos, Mohamed, , and Vedaldi]{cimpoi14describing}
M. Cimpoi, S. Maji, I. Kokkinos, S. Mohamed, , and A. Vedaldi.
\newblock Describing textures in the wild.
\newblock In \emph{CVPR}, 2014.

\bibitem[Dehghani et~al.(2023)Dehghani, Djolonga, Mustafa, Padlewski, Heek, Gilmer, Steiner, Caron, Geirhos, Alabdulmohsin, et~al.]{dehghani2023scaling}
Mostafa Dehghani, Josip Djolonga, Basil Mustafa, Piotr Padlewski, Jonathan Heek, Justin Gilmer, Andreas~Peter Steiner, Mathilde Caron, Robert Geirhos, Ibrahim Alabdulmohsin, et~al.
\newblock Scaling vision transformers to 22 billion parameters.
\newblock In \emph{ICML}, 2023.

\bibitem[Deng et~al.(2009)Deng, Dong, Socher, Li, Li, and Fei-Fei]{ImageNet}
Jia Deng, Wei Dong, Richard Socher, Li-Jia Li, Kai Li, and Li Fei-Fei.
\newblock Imagenet: A large-scale hierarchical image database.
\newblock In \emph{CVPR}, pages 248--255, 2009.

\bibitem[Deshpande et~al.(2019)Deshpande, Hu, Sun, Pyrros, Siddiqui, Koyejo, Zhao, Forsyth, and Schwing]{Deshpande2019}
Ishani Deshpande, Yuan-Ting Hu, Ruoyu Sun, Ayis Pyrros, Nasir Siddiqui, Oluwasanmi Koyejo, Zhizhen Zhao, David~Alexander Forsyth, and Alexander~G. Schwing.
\newblock Max-sliced wasserstein distance and its use for gans.
\newblock \emph{CVPR}, 2019.

\bibitem[Diao et~al.(2023)Diao, Wang, Zhang, Yang, Ding, and Tarokh]{diaopruning}
Enmao Diao, Ganghua Wang, Jiawei Zhang, Yuhong Yang, Jie Ding, and Vahid Tarokh.
\newblock Pruning deep neural networks from a sparsity perspective.
\newblock In \emph{ICLR}, 2023.

\bibitem[Dror et~al.(2022)Dror, Zehngut, Raviv, Artyomov, Vitek, and Jevnisek]{dror2021layer}
Amir~Ben Dror, Niv Zehngut, Avraham Raviv, Evgeny Artyomov, Ran Vitek, and Roy Jevnisek.
\newblock Layer folding: Neural network depth reduction using activation linearization.
\newblock \emph{BMVC}, 2022.

\bibitem[Fang et~al.(2023)Fang, Ma, Song, Mi, and Wang]{fang2023depgraph}
Gongfan Fang, Xinyin Ma, Mingli Song, Michael~Bi Mi, and Xinchao Wang.
\newblock Depgraph: Towards any structural pruning.
\newblock In \emph{CVPR}, 2023.

\bibitem[Flamary et~al.(2021)Flamary, Courty, Gramfort, Alaya, Boisbunon, Chambon, Chapel, Corenflos, Fatras, Fournier, Gautheron, Gayraud, Janati, Rakotomamonjy, Redko, Rolet, Schutz, Seguy, Sutherland, Tavenard, Tong, and Vayer]{pot}
Rémi Flamary, Nicolas Courty, Alexandre Gramfort, Mokhtar~Z. Alaya, Aurélie Boisbunon, Stanislas Chambon, Laetitia Chapel, Adrien Corenflos, Kilian Fatras, Nemo Fournier, Léo Gautheron, Nathalie~T.H. Gayraud, Hicham Janati, Alain Rakotomamonjy, Ievgen Redko, Antoine Rolet, Antony Schutz, Vivien Seguy, Danica~J. Sutherland, Romain Tavenard, Alexander Tong, and Titouan Vayer.
\newblock Pot: Python optimal transport.
\newblock \emph{Journal of Machine Learning Research (JMLR)}, 2021.

\bibitem[Gale et~al.(2019)Gale, Elsen, and Hooker]{Gale_Magnitude}
Trevor Gale, Erich Elsen, and Sara Hooker.
\newblock The state of sparsity in deep neural networks.
\newblock \emph{arXiv preprint arXiv:1902.09574}, 2019.

\bibitem[Gholami et~al.(2022)Gholami, Kim, Dong, Yao, Mahoney, and Keutzer]{gholami2022survey}
Amir Gholami, Sehoon Kim, Zhen Dong, Zhewei Yao, Michael~W Mahoney, and Kurt Keutzer.
\newblock A survey of quantization methods for efficient neural network inference.
\newblock In \emph{Low-Power Computer Vision}, pages 291--326. Chapman and Hall/CRC, 2022.

\bibitem[Gromov et~al.(2025)Gromov, Tirumala, Shapourian, Glorioso, and Roberts]{gromov2024unreasonable}
Andrey Gromov, Kushal Tirumala, Hassan Shapourian, Paolo Glorioso, and Daniel~A Roberts.
\newblock The unreasonable ineffectiveness of the deeper layers.
\newblock \emph{ICLR}, 2025.

\bibitem[Gulrajani and Lopez-Paz(2020)]{gulrajani2020search}
Ishaan Gulrajani and David Lopez-Paz.
\newblock In search of lost domain generalization.
\newblock In \emph{ICLR}, 2020.

\bibitem[Han et~al.(2015)Han, Pool, Tran, and Dally]{han2015learning}
Song Han, Jeff Pool, John Tran, and William Dally.
\newblock Learning both weights and connections for efficient neural network.
\newblock \emph{NeurIPS}, 2015.

\bibitem[He and Xiao(2023)]{he2023structured}
Yang He and Lingao Xiao.
\newblock Structured pruning for deep convolutional neural networks: A survey.
\newblock \emph{IEEE PAMI}, 2023.

\bibitem[Hestness et~al.(2017)Hestness, Narang, Ardalani, Diamos, Jun, Kianinejad, Patwary, Yang, and Zhou]{hestness2017deep}
Joel Hestness, Sharan Narang, Newsha Ardalani, Gregory Diamos, Heewoo Jun, Hassan Kianinejad, Md~Mostofa~Ali Patwary, Yang Yang, and Yanqi Zhou.
\newblock Deep learning scaling is predictable, empirically.
\newblock \emph{arXiv preprint arXiv:1712.00409}, 2017.

\bibitem[Heusel et~al.(2017)Heusel, Ramsauer, Unterthiner, Nessler, and Hochreiter]{heusel2017gans}
Martin Heusel, Hubert Ramsauer, Thomas Unterthiner, Bernhard Nessler, and Sepp Hochreiter.
\newblock Gans trained by a two time-scale update rule converge to a local nash equilibrium.
\newblock \emph{NeurIPS}, 2017.

\bibitem[Hinton(2015)]{hinton2015distilling}
Geoffrey Hinton.
\newblock Distilling the knowledge in a neural network.
\newblock \emph{arXiv preprint arXiv:1503.02531}, 2015.

\bibitem[Jia et~al.(2021)Jia, Yang, Xia, Chen, Parekh, Pham, Le, Sung, Li, and Duerig]{jia2021scaling}
Chao Jia, Yinfei Yang, Ye Xia, Yi-Ting Chen, Zarana Parekh, Hieu Pham, Quoc Le, Yun-Hsuan Sung, Zhen Li, and Tom Duerig.
\newblock Scaling up visual and vision-language representation learning with noisy text supervision.
\newblock In \emph{ICML}, 2021.

\bibitem[Kandasamy et~al.(2018)Kandasamy, Neiswanger, Schneider, Poczos, and Xing]{NEURIPS2018_f33ba15e}
Kirthevasan Kandasamy, Willie Neiswanger, Jeff Schneider, Barnabas Poczos, and Eric~P Xing.
\newblock Neural architecture search with bayesian optimisation and optimal transport.
\newblock In \emph{NeurIPS}, 2018.

\bibitem[Karkar et~al.(2020)Karkar, Ayed, de~Bezenac, and Gallinari]{leastact}
Skander Karkar, Ibrahim Ayed, Emmanuel de Bezenac, and Patrick Gallinari.
\newblock {A Principle of Least Action for the Training of Neural Networks}.
\newblock In \emph{ECML PKDD}, 2020.

\bibitem[Kirillov et~al.(2023)Kirillov, Mintun, Ravi, Mao, Rolland, Gustafson, Xiao, Whitehead, Berg, Lo, et~al.]{kirillov2023segment}
Alexander Kirillov, Eric Mintun, Nikhila Ravi, Hanzi Mao, Chloe Rolland, Laura Gustafson, Tete Xiao, Spencer Whitehead, Alexander~C Berg, Wan-Yen Lo, et~al.
\newblock Segment anything.
\newblock In \emph{ICCV}, 2023.

\bibitem[Krizhevsky et~al.(2009)Krizhevsky, Hinton, et~al.]{krizhevsky2009learning}
Alex Krizhevsky, Geoffrey Hinton, et~al.
\newblock Learning multiple layers of features from tiny images, 2009.

\bibitem[Le and Yang(2015)]{le2015tiny}
Ya Le and Xuan Yang.
\newblock Tiny imagenet visual recognition challenge.
\newblock \emph{CS 231N}, 7\penalty0 (7):\penalty0 3, 2015.

\bibitem[Lee et~al.(2019)Lee, Ajanthan, and Torr]{lee2018snip}
Namhoon Lee, Thalaiyasingam Ajanthan, and Philip Torr.
\newblock Snip: Single-shot network pruning based on connection sensitivity.
\newblock In \emph{ICLR}, 2019.

\bibitem[Li et~al.(2019)Li, Haque, Anil, Lucas, Grosse, and Jacobsen]{Li2019PreventingGA}
Qiyang Li, Saminul Haque, Cem Anil, James Lucas, Roger~Baker Grosse, and Joern-Henrik Jacobsen.
\newblock Preventing gradient attenuation in lipschitz constrained convolutional networks.
\newblock \emph{NeurIPS}, 2019.

\bibitem[Liao et~al.(2023)Liao, Qu{\'e}tu, Nguyen, and Tartaglione]{liao2023can}
Zhu Liao, Victor Qu{\'e}tu, Van-Tam Nguyen, and Enzo Tartaglione.
\newblock Can unstructured pruning reduce the depth in deep neural networks?
\newblock In \emph{ICCV}, 2023.

\bibitem[Liao et~al.(2024)Liao, Qu{\'e}tu, Nguyen, and Tartaglione]{liao2024nepenthe}
Zhu Liao, Victor Qu{\'e}tu, Van-Tam Nguyen, and Enzo Tartaglione.
\newblock Nepenthe: Entropy-based pruning as a neural network depth's reducer.
\newblock \emph{arXiv preprint arXiv:2404.16890}, 2024.

\bibitem[Liebenwein et~al.(2021)Liebenwein, Baykal, Carter, Gifford, and Rus]{liebenwein2021lost}
Lucas Liebenwein, Cenk Baykal, Brandon Carter, David Gifford, and Daniela Rus.
\newblock Lost in pruning: The effects of pruning neural networks beyond test accuracy.
\newblock \emph{Proceedings of Machine Learning and Systems}, 3:\penalty0 93--138, 2021.

\bibitem[Liu et~al.(2021)Liu, Zhuang, Zhuang, Guo, Huang, Zhu, and Tan]{liu2021discrimination}
Jing Liu, Bohan Zhuang, Zhuangwei Zhuang, Yong Guo, Junzhou Huang, Jinhui Zhu, and Mingkui Tan.
\newblock Discrimination-aware network pruning for deep model compression.
\newblock \emph{TPAMI}, 2021.

\bibitem[Lohit and Jones(2022)]{lohit2022model}
Suhas Lohit and Michael Jones.
\newblock Model compression using optimal transport.
\newblock In \emph{WACV}, 2022.

\bibitem[Louizos et~al.(2018)Louizos, Welling, and Kingma]{louizos2018learning}
Christos Louizos, Max Welling, and Diederik~P. Kingma.
\newblock Learning sparse neural networks through $l_0$ regularization.
\newblock In \emph{ICLR}, 2018.

\bibitem[Maji et~al.(2013)Maji, Kannala, Rahtu, Blaschko, and Vedaldi]{maji13fine-grained}
S. Maji, J. Kannala, E. Rahtu, M. Blaschko, and A. Vedaldi.
\newblock Fine-grained visual classification of aircraft, 2013.

\bibitem[Mehmeti-G{\"o}pel and Disselhoff(2023)]{mehmeti2023nonlinear}
Christian HX~Ali Mehmeti-G{\"o}pel and Jan Disselhoff.
\newblock Nonlinear advantage: trained networks might not be as complex as you think.
\newblock In \emph{ICML}, 2023.

\bibitem[Nadjahi et~al.(2020)Nadjahi, Durmus, Chizat, Kolouri, Shahrampour, and Simsekli]{Nadjahi2020StatisticalAT}
Kimia Nadjahi, Alain Durmus, L{\'e}na{\"i}c Chizat, Soheil Kolouri, Shahin Shahrampour, and Umut Simsekli.
\newblock Statistical and topological properties of sliced probability divergences.
\newblock \emph{NeurIPS}, 2020.

\bibitem[Nadjahi et~al.(2021)Nadjahi, Durmus, Jacob, Badeau, and Simsekli]{nadjahi2021fast}
Kimia Nadjahi, Alain Durmus, Pierre Jacob, Roland Badeau, and Umut Simsekli.
\newblock Fast approximation of the sliced-wasserstein distance using concentration of random projections.
\newblock In \emph{NeurIPS}, 2021.

\bibitem[Nguyen et~al.(2021)Nguyen, Le, Yamada, and Osborne]{nguyen2021optimal}
Vu Nguyen, Tam Le, Makoto Yamada, and Michael~A Osborne.
\newblock Optimal transport kernels for sequential and parallel neural architecture search.
\newblock In \emph{ICML}, 2021.

\bibitem[Nilsback and Zisserman(2008)]{Nilsback08}
Maria-Elena Nilsback and Andrew Zisserman.
\newblock Automated flower classification over a large number of classes.
\newblock In \emph{Indian Conference on Computer Vision, Graphics and Image Processing}, 2008.

\bibitem[Parmar et~al.(2022)Parmar, Zhang, and Zhu]{parmar2022aliased}
Gaurav Parmar, Richard Zhang, and Jun-Yan Zhu.
\newblock On aliased resizing and surprising subtleties in gan evaluation.
\newblock In \emph{CVPR}, 2022.

\bibitem[Peebles and Xie(2023)]{peebles2023scalable}
William Peebles and Saining Xie.
\newblock Scalable diffusion models with transformers.
\newblock In \emph{ICCV}, 2023.

\bibitem[Peyr{\'e} et~al.(2019)Peyr{\'e}, Cuturi, et~al.]{peyre2019computational}
Gabriel Peyr{\'e}, Marco Cuturi, et~al.
\newblock Computational optimal transport: With applications to data science.
\newblock \emph{Foundations and Trends{\textregistered} in Machine Learning}, 11\penalty0 (5-6):\penalty0 355--607, 2019.

\bibitem[Pilo et~al.(2025)Pilo, Hezbri, e~Ferreira, Qu{\'e}tu, and Tartaglione]{unknown}
Giommaria Pilo, Nour Hezbri, Andr{\'e}~Pereira e Ferreira, Victor Qu{\'e}tu, and Enzo Tartaglione.
\newblock Layerfold: A python library to reduce the depth of neural networks.
\newblock \emph{SoftwareX}, 2025.

\bibitem[Qu{\'e}tu and Tartaglione(2024)]{quetu2024dsd2}
Victor Qu{\'e}tu and Enzo Tartaglione.
\newblock Dsd$^2$: Can we dodge sparse double descent and compress the neural network worry-free?
\newblock In \emph{AAAI}, 2024.

\bibitem[Qu{\'e}tu et~al.(2024)Qu{\'e}tu, Liao, and Tartaglione]{quetu2024simpler}
Victor Qu{\'e}tu, Zhu Liao, and Enzo Tartaglione.
\newblock The simpler the better: An entropy-based importance metric to reduce neural networks’ depth.
\newblock In \emph{ECML PKDD}, 2024.

\bibitem[Radford et~al.(2021)Radford, Kim, Hallacy, Ramesh, Goh, Agarwal, Sastry, Askell, Mishkin, Clark, et~al.]{radford2021learning}
Alec Radford, Jong~Wook Kim, Chris Hallacy, Aditya Ramesh, Gabriel Goh, Sandhini Agarwal, Girish Sastry, Amanda Askell, Pamela Mishkin, Jack Clark, et~al.
\newblock Learning transferable visual models from natural language supervision.
\newblock In \emph{ICML}, 2021.

\bibitem[Ramesh et~al.(2022)Ramesh, Dhariwal, Nichol, Chu, and Chen]{ramesh2022hierarchical}
Aditya Ramesh, Prafulla Dhariwal, Alex Nichol, Casey Chu, and Mark Chen.
\newblock Hierarchical text-conditional image generation with clip latents.
\newblock \emph{arXiv preprint arXiv:2204.06125}, 1\penalty0 (2):\penalty0 3, 2022.

\bibitem[Rombach et~al.(2022)Rombach, Blattmann, Lorenz, Esser, and Ommer]{rombach2022high}
Robin Rombach, Andreas Blattmann, Dominik Lorenz, Patrick Esser, and Bj{\"o}rn Ommer.
\newblock High-resolution image synthesis with latent diffusion models.
\newblock In \emph{CVPR}, 2022.

\bibitem[Rosenfeld(2021)]{rosenfeld2021scaling}
Jonathan~S Rosenfeld.
\newblock Scaling laws for deep learning.
\newblock \emph{arXiv preprint arXiv:2108.07686}, 2021.

\bibitem[Seyfarth et~al.(2024)Seyfarth, Dar, and Engelhardt]{seyfarth2024latent}
Marvin Seyfarth, Salman Ul~Hassan Dar, and Sandy Engelhardt.
\newblock Latent pollution model: The hidden carbon footprint in 3d image synthesis.
\newblock In \emph{International Workshop on Simulation and Synthesis in Medical Imaging}, 2024.

\bibitem[Si et~al.(2020)Si, Blanchet, Ghosh, and Squillante]{curse_dim}
Nian Si, Jose Blanchet, Soumyadip Ghosh, and Mark Squillante.
\newblock Quantifying the empirical wasserstein distance to a set of measures: Beating the curse of dimensionality.
\newblock In \emph{NeurIPS}, 2020.

\bibitem[Tartaglione et~al.(2021)Tartaglione, Bragagnolo, Odierna, Fiandrotti, and Grangetto]{tartaglione2021serene}
Enzo Tartaglione, Andrea Bragagnolo, Francesco Odierna, Attilio Fiandrotti, and Marco Grangetto.
\newblock Serene: Sensitivity-based regularization of neurons for structured sparsity in neural networks.
\newblock \emph{IEEE Transactions on Neural Networks and Learning Systems}, 33\penalty0 (12):\penalty0 7237--7250, 2021.

\bibitem[Tartaglione et~al.(2022)Tartaglione, Bragagnolo, Fiandrotti, and Grangetto]{tartaglione2022loss}
Enzo Tartaglione, Andrea Bragagnolo, Attilio Fiandrotti, and Marco Grangetto.
\newblock Loss-based sensitivity regularization: towards deep sparse neural networks.
\newblock \emph{Neural Networks}, 2022.

\bibitem[Titouan et~al.(2019)Titouan, Flamary, Courty, Tavenard, and Chapel]{NEURIPS2019_a9cc6694}
Vayer Titouan, R\'{e}mi Flamary, Nicolas Courty, Romain Tavenard, and Laetitia Chapel.
\newblock Sliced gromov-wasserstein.
\newblock In \emph{NeurIPS}, 2019.

\bibitem[Villani et~al.(2009)]{villani2009optimal}
C{\'e}dric Villani et~al.
\newblock \emph{Optimal transport: old and new}.
\newblock Springer, 2009.

\bibitem[Wu et~al.(2022)Wu, Raghavendra, Gupta, Acun, Ardalani, Maeng, Chang, Aga, Huang, Bai, Gschwind, Gupta, Ott, Melnikov, Candido, Brooks, Chauhan, Lee, Lee, Akyildiz, Balandat, Spisak, Jain, Rabbat, and Hazelwood]{MLSYS2022_462211f6}
Carole-Jean Wu, Ramya Raghavendra, Udit Gupta, Bilge Acun, Newsha Ardalani, Kiwan Maeng, Gloria Chang, Fiona Aga, Jinshi Huang, Charles Bai, Michael Gschwind, Anurag Gupta, Myle Ott, Anastasia Melnikov, Salvatore Candido, David Brooks, Geeta Chauhan, Benjamin Lee, Hsien-Hsin Lee, Bugra Akyildiz, Maximilian Balandat, Joe Spisak, Ravi Jain, Mike Rabbat, and Kim Hazelwood.
\newblock Sustainable ai: Environmental implications, challenges and opportunities.
\newblock In \emph{Proceedings of Machine Learning and Systems}, pages 795--813, 2022.

\bibitem[Yang et~al.(2023)Yang, Liu, and Xu]{yang2023hotnas}
Jiechao Yang, Yong Liu, and Hongteng Xu.
\newblock Hotnas: Hierarchical optimal transport for neural architecture search.
\newblock In \emph{CVPR}, 2023.

\bibitem[Zhu and Gupta(2018)]{h.2018to}
Michael~H. Zhu and Suyog Gupta.
\newblock To prune, or not to prune: Exploring the efficacy of pruning for model compression.
\newblock In \emph{ICLR}, 2018.

\bibitem[Zhu et~al.(2021)Zhu, DING, Zhou, Li, You, Sulam, and Qu]{zhu2021a}
Zhihui Zhu, Tianyu DING, Jinxin Zhou, Xiao Li, Chong You, Jeremias Sulam, and Qing Qu.
\newblock A geometric analysis of neural collapse with unconstrained features.
\newblock In \emph{NeurIPS}, 2021.

\end{thebibliography}
}

\clearpage
\appendix
\setcounter{page}{1}
\maketitlesupplementary

\section{Gradients of the regularizer }
Herein, we provide further details on the regularizer, namely, by deriving its gradients.

Namely,~$\forall k_0 \in \llbracket K \rrbracket $, 
\begin{equation}
    \mathcal{R}_{k_0}=\sqrt{\frac{1}{N} \sum_{i=1}^N (\theta^T\boldsymbol{x}_{k_0,i} - \theta^TT_{k_0}(\boldsymbol{x}_{k_0,i}; w_{k_0}))^2} 
\end{equation}
In our setting, we are also assuming that ~$ T_{k_0}(\boldsymbol{x}_{k_0,i};w_{k_0})= \boldsymbol{x}_{k_0,i} + f_{k_0}(\boldsymbol{x}_{k_0,i}; w_{k_0})$. Hence, we make use of this expression to derive the following analytical expression for ~$\frac{\partial \mathcal{R}_{k_0}}{\partial w_{k_0}}$. 

\begin{equation}
    \frac{\partial \mathcal{R}_{k_0}}{\partial w_{k_0}}=\frac{1}{N \mathcal{R}_{k_0}} \sum_{i=1}^N (\theta^Tf_{k_0}(\boldsymbol{x}_{k_0,i}; w_{k_0})) \frac{\partial \theta^Tf_{k_0}(\boldsymbol{x}_{k_0,i}; w_{k_0})}{\partial w_{k_0}}   
\end{equation}

\begin{equation}
 \Rightarrow    \frac{\partial \mathcal{R} }{\partial w_{k_0}}=\frac{1}{K} \sum_{k=k_0}^K\frac{\partial \mathcal{R}_{k}}{\partial w_{k_0}}  
\end{equation}

Computing the Wasserstein distance requires a sorting oracle, even in one dimension. 
In practice, backpropagation works through subgradients of the sorting operation. While argsort is non-differentiable, by leveraging automatic differentiation with PyTorch, in the POT library~\cite{pot} gradients flow through the \textit{take\_along\_axis} operation, treating sorting indices as constants during backpropagation. This is sufficient because the Wasserstein distance remains well-defined even at points where the sorting order changes.

\section{Correlation between Wasserstein distance and performance degradation}

In Tab.~\ref{tab:indicator}, we analyze the correlation between the Wasserstein distance for each considered block (B1, B2, B3, B4), and performance degradation for a ResNet-18 on CIFAR-10. The higher $\lambda$, the lower the distances, thus the more likely the block has a function close to the identity, and therefore the higher the performance since removing a block close to identity does not lead to any changes.
Indeed, if the Wasserstein distance is zero, by definition the output matches the input and therefore the whole block encodes the identity function. The higher the distance is, the larger the perturbation introduced will be, increasing the risk of performance loss.
Consequently, as $\lambda$ increases, performance improves since removing a block functioning close to identity has minimal impact on the model's behavior.

\begin{table}[t]
    \centering
    \resizebox{\columnwidth}{!}{%
    \begin{tabular}{c c c c c c c}
    \toprule
        \bf $\lambda$ & \bf B1 & \bf B2 & \bf B3 & \bf B4 & \bf Mean & \bf top-1 \\ 
    \midrule 
        0.01 & 0.210 & 0.126 & 0.104 & 0.044 & 0.121 & 73.04 \\ 
    
        0.1 & 0.068 & 0.074 & 0.036 & 0.016 & 0.048 & 80.23 \\
    
        1 & 4.63e-4 & 0.0246 & 5.80e-3 & 2.59e-4 & 7.78e-3 & 89.01 \\
    
        5 & 4.41e-4 & 7.93e-3 & 4.35e-4 & 1.34e-4 & 2.23e-3 & 90.99 \\
    \bottomrule
    
    \end{tabular}%
    }
    \caption{Correlation between the Wasserstein distance of each block (B1, B2, B3, B4) and the performance for a ResNet-18 on CIFAR-10.}
    \label{tab:indicator}
\end{table}
\section{Relationship between critical path length and practical resource consumption}

\begin{figure}[ht]
    \centering
    \includegraphics[width=\columnwidth]{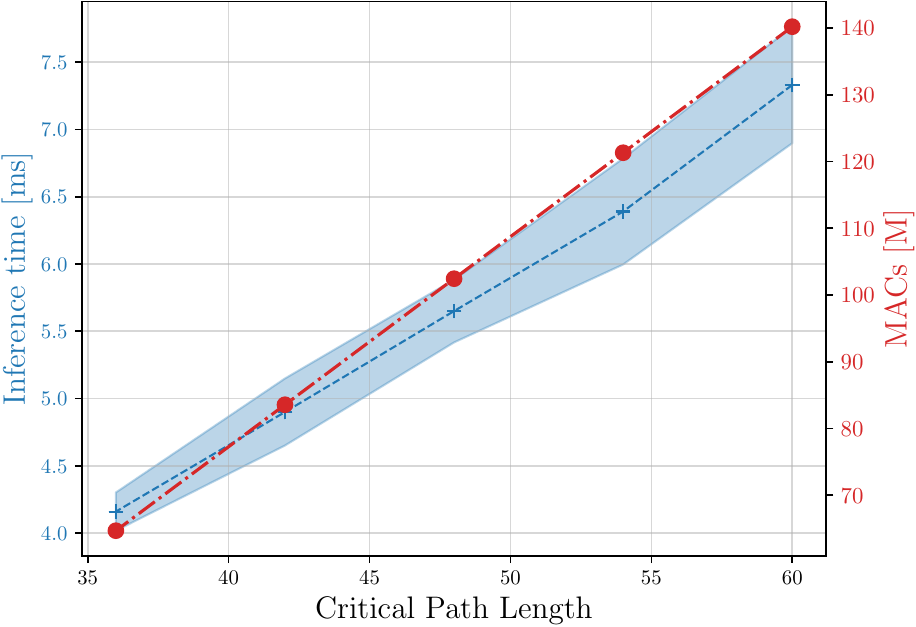}
    \caption{Relationship between Critical Path Length (CPL), inference time, and Multiply-Accumulate Operations (MACs) for a ResNet-18 model on the CIFAR-10 dataset. As the CPL decreases, both the inference time and the number of MACs decrease, indicating improved computational efficiency and faster inference speeds.}
    \label{fig:CPL}
\end{figure}

In this section, we show that optimizing the critical path length can lead to significant improvements in both inference speed and computational efficiency. Indeed, Fig.~\ref{fig:CPL} demonstrates the relationship between critical path length, inference time (measured on a NVIDIA A4500) and computational complexity (MACs) for a ResNet-18 on CIFAR-10.

This analysis highlights a trade-off between computational complexity and inference speed. Reducing the critical path length leads to both faster inference times and fewer computational operations: shorter critical paths are more efficient in terms of both time and computational resources.
\section{Practical Benefits}
\label{appendix:practical}
Following the analysis conducted in Sec.~\ref{subsec:pratical}, in this subsection, we showcase the practical benefits of our approach in terms of efficiency as well as inference time.

To clarify the role of block eligibility in our method, Tab.~\ref{tab:blocks} reports, for each tested backbone: the total number of blocks ($\#$ Blocks), the number of blocks satisfying the equal-shape criterion and thus eligible for MSW scoring ($\#$ MSW), the number of blocks ultimately pruned ($\#$ Pruned), along with the resulting percentage reductions in latency and MACs, and the resulting Top-1 accuracy on CIFAR-10. Tab.~\ref{tab:blocks} provides a detailed quantification of block eligibility and its impact on both efficiency and performance. For the cases where this assumption does not hold, we describe a workaround in Sec.~\ref{appendix:mismatch}. 
\begin{table}[!h]
\centering
\resizebox{0.95\columnwidth}{!}{%
    \centering
    \begin{tabular}{c c c c c c c}
    \toprule
        \textbf{Backbone} & \textbf{\# Blocks} & \textbf{\# MSW} & \textbf{\# Pruned} & \textbf{Latency} & \textbf{MACs} & \textbf{Top-1} \\ 
    \midrule
        ResNet-18 & 8 & 4 & 4 & -39.50 \% & -53.86 \% & 90.99 \\
    \midrule
        Swin-T & 12 & 12 & 3 & -38.48 \% & -61.55 \% & 89.47 \\
    \midrule
        MobileNetv2 & 17 & 12 & 10 & -23.66 \% & -3.81 \% & 87.25 \\
    \bottomrule
    \end{tabular}
}
\caption{Block eligibility and impact of pruning on CIFAR-10.} 
\label{tab:blocks}
\end{table}

Tab.\ref{tab:C10-MNv2_ablation} shows the test performance, MACs, and inference time on an NVIDIA A4500, as well as the training time for a MobileNetv2 trained on CIFAR-10 for all the considered approaches.
\begin{table}[!h]
    \centering
    \resizebox{\columnwidth}{!}{%
    \begin{tabular}{c|c|c|c|c}
    \toprule
        \textbf{Approach} & \textbf{top-1 [\%]} & \textbf{MACs [M]} & \textbf{Inference time [ms]} & \textbf{Time} \\
    \toprule 
        Original & 93.50 & 87.98 & 13.57 $\pm$ 0.82 & 112' \\
    \midrule
        Layer Folding & 86.56 &  87.98 & 20.29 $\pm$ 0.19 & 529' \\
        EGP & 9.70 & 87.98 & 13.38 $\pm$ 0.18 & 732' \\
        NEPENTHE & 86.75 & 87.98 & 13.29 $\pm$ 0.24 & 3165' \\
        EASIER & 87.19 & 87.98 & 13.22 $\pm$ 0.54 & 3514' \\
    \midrule
        LaCoOT & \bf 87.25 & \bf 84.63 & \bf 10.36 $\pm$ 0.60 & \bf 132' \\  
    \bottomrule
    \end{tabular}
    }
    \caption{Test performance (top-1), MACs, inference time on a NVIDIA A4500 and training time for MobileNetv2 trained on CIFAR-10. Original refers to the trained model without layer deletion. The best results between Layer Folding, EGP, NEPENTHE, EASIER and LaCoOT are in \textbf{bold}.}
    \label{tab:C10-MNv2_ablation}
\end{table}

In this setup, while Layer Folding and EGP showcase performance drop at high critical path length, we achieve comparable performance as EASIER or NEPENTHE for the same critical path length in 20$\times$ less time, with real practical benefits since the inference time and the MACs are reduced.

Tab.~\ref{tab:C10-S_ablation} shows the test performance, MACs, and inference time on an NVIDIA A4500, as well as the training time for a Swin-T trained on CIFAR-10 for all the considered approaches.
In this setup, since the fusion of two linear layers is straightforward, we merge the layers for the baseline methods when the non-linearity in between has been removed.

\begin{table}[!h]
    \centering
    \resizebox{\columnwidth}{!}{%
    \begin{tabular}{c|c|c|c|c}
    \toprule
        \textbf{Approach} & \textbf{top-1 [\%]} & \textbf{MACs [M]} & \textbf{Inference time [ms]} & \textbf{Time} \\
    \toprule 
        Original & 91.67 & 518.94 & 13.54 $\pm$ 0.32 & 113' \\
    \midrule
        Layer Folding & 85.73 &  510.80 & 14.89 $\pm$ 0.11 & 383' \\
        EGP & 92.01 & 514.95 & 13.51 $\pm$ 0.17 & 228' \\ 
        NEPENTHE & \bf 92.29 & 510.82 & 13.24 $\pm$ 0.26 & 688' \\
        EASIER & 91.25 & 494.28 & 11.04 $\pm$ 0.15 & 803' \\
    \midrule
        LaCoOT & 89.47 & \bf 199.54 & \bf 8.33 $\pm$ 0.02 & \bf 135' \\ 
    \bottomrule
    \end{tabular}
    }
    \caption{Test performance (top-1), MACs, inference time on a NVIDIA A4500 and training time for Swin-T trained on CIFAR-10. Original refers to the trained model without layer deletion. The best results between Layer Folding, EGP, NEPENTHE, EASIER and LaCoOT are in \textbf{bold}.}
    \label{tab:C10-S_ablation}
\end{table}

In this setup, we can observe that the other methods reduce the number of MACs. However, there is little (if any) benefit in practice: the inference time is not reduced.  Indeed, these methods focus solely on removing non-linearities, unlike our method, which removes complete blocks.
On the other hand, although a slight loss of performance is noticeable, our method LaCoOT considerably reduces the number of MACs and decreases the inference time by more than 35\%, while being far more efficient at training time than its competitors. 
\section{Extension of LaCoOT to layers with mismatched dimensionalities}
\label{appendix:mismatch}
LaCoOT was primilary designed to operate on layers where the Max-Sliced Wasserstein distance can be computed directly. 
This distance requires matching dimensions between distributions, which prevents a direct calculation of this distance for layers with different input and output dimensions. 
Nevertheless, we propose in this section to address this issue by studying the case of a 3x3 convolutional layer inside a ResNet-18. 

Our goal here is to remove this specific layer in the network. However, directly removing the layer would result in a mismatch in both spatial resolution and channel dimensionality, disrupting the flow of activations through the network.
To mitigate this, we introduce an alternative transformation that preserves the overall network structure while ensuring compatibility with subsequent layers.
Specifically, we replace the 3×3 convolutional layer with a combination of a spatial downsampling operation and a 1×1 convolution. The downsampling is achieved using an average pooling layer (AvgPool2d) with a 2×2 kernel and a stride of 2, which reduces the spatial resolution. The 1×1 convolution then adjusts the number of output channels to match the expected input dimensions of the following layers. 
To ensure both configurations could be trained simultaneously, we implemented a dual-path approach within the modified block, where both the original and new transformations coexisted.
During training, the introduced path with the 1x1 convolution is only trained using the Max-Sliced Wasserstein distance, computed between the output distribution of the 1x1 convolution and the output distribution of the original 3x3 convolution.
Post-training, the original 3x3 convolution is discarded, and replaced by the average pooling and the newly trained 1x1 convolution. 

Following the same training policy detailed in Sec.~\ref{appendix:details}, a ResNet-18 with the introduced transformation is trained on CIFAR-10.
On the one hand, with $\lambda=0$, the resulting network with the proposed transformation loses 0.98\% performance compared to its full version. On the other hand, with $\lambda=0.1$, the resulting network achieves comparable performance with only a 0.18\% performance loss compared to its full version, highlighting the effectiveness of our method in this case. 

To conclude, by incorporating this modified structure into the ResNet-18 architecture, we enable a seamless integration of LaCoOT which addresses the case of layers with mismatched dimensions.
\section{Additional Results on Image Classification Setups}
\label{appendix:more_results}
\begin{figure*}[t]
    \centering
    \includegraphics[width=\linewidth]{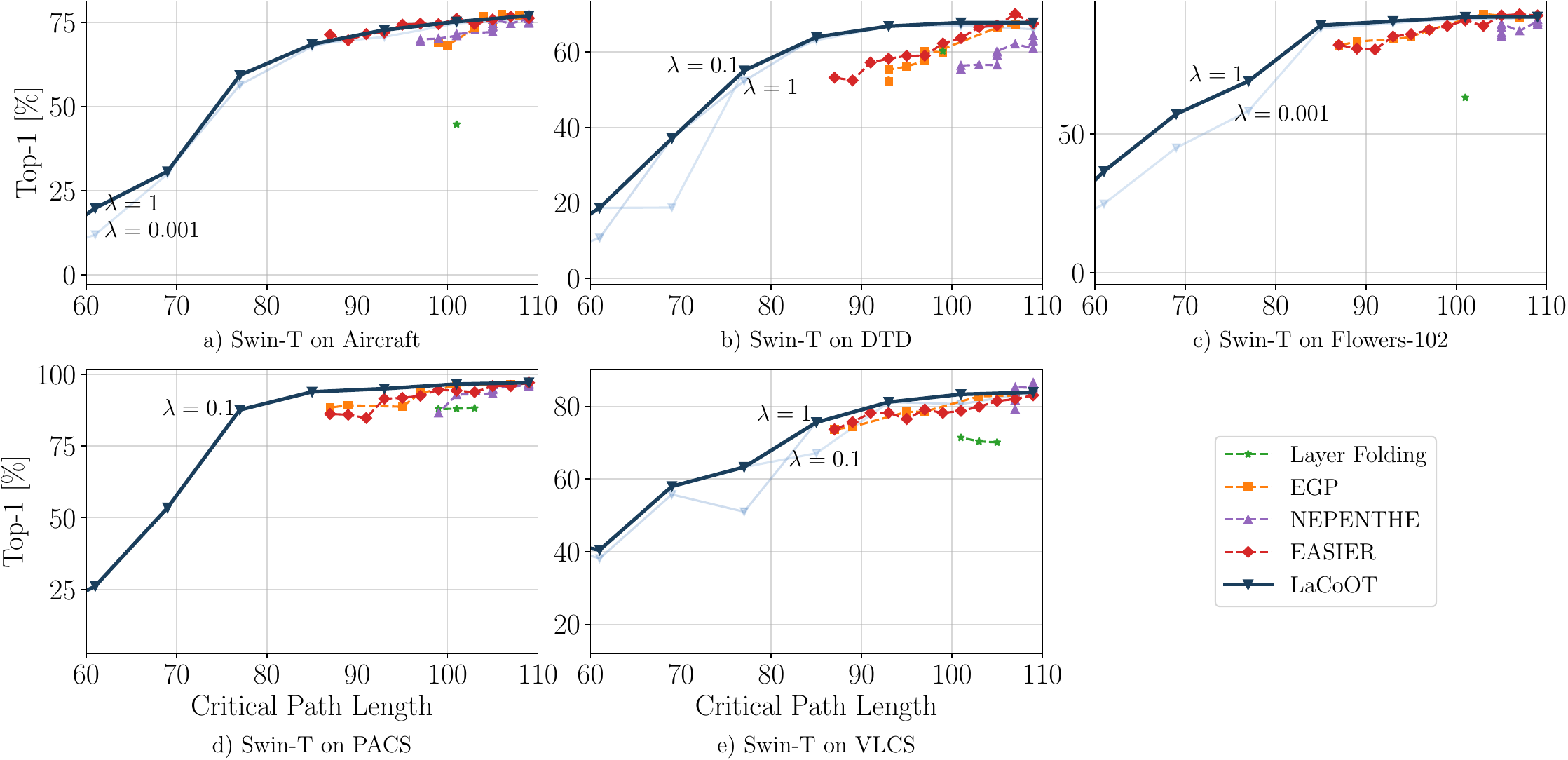}
    \caption{Test performance (Top-1 [\%]) in function of the Critical Path Length for Swin-T trained on Aircraft (a), DTD (b), Flowers-102 (c), PACS (d) and VLCS (e). For each setup, we showcase the results achieved by LaCoOT for different values of $\lambda$, forming in \textbf{\textcolor{dark_blue}{dark blue}} the pareto frontier of our technique. \emph{Top left corner is the best.}}
    \label{fig:Swin-T_appendix}
\end{figure*}

To complete the comparisons carried out in Sec.~\ref{sec:results}, we compare in this section the effectiveness of LaCoOT with respect to other baselines methods on Swin-T trained on PACS, VLCS, Aircraft, Flowers-102 and DTD in Fig~\ref{fig:Swin-T_appendix}. Indeed, as demonstrated in the previous section (Sec.~\ref{appendix:practical}), Swin-T is the only architecture where competing methods can lead to practical benefits, since the fusion of two consecutive linear layers is straightforward.

In most setups, we can observe the effectiveness of LaCoOT. Indeed, for short critical path length, LaCoOT is the method performing overall the best, achieving a new Pareto Frontier when $\lambda$ is increasing.
In some cases like DTD, LaCoOT outperforms current methods by 10\% for the same critical path length. Additionally, looking at longer critical path lengths, we can observe that when $\lambda$ is decreasing, LaCoOT achieves comparable results to other baseline methods.
Overall, since our method focuses on removing blocks, shorter critical path lengths can be achieved even though the performance drops dramatically. 
For very short critical path length (around 60), applying a healing policy or finetuning to the model could help recovering performance. However, we leave this aspect to future work.
\section{Ranking with the Lipschitz constant}
While Fig.~\ref{fig:main_res} and \ref{fig:DiT-FID} emphasize critical path length, we also report results using standard metrics (MACs, inference time, and training time) in Tab.~\ref{tab:C10-R18_ablation}, \ref{tab:C10-MNv2_ablation} and \ref{tab:C10-S_ablation}, which confirm the advantage of LaCoOT across multiple resource and performance dimensions. Furthermore, we show here in Tab.~\ref{tab:Lipschitz} that the ranking of the methods is preserved by displaying the Lipschitz constants across all blocks (B1–B4) of ResNet-18 on CIFAR-10. The global Lipschitz constant is upper bounded by the product of each block's Lipschitz constant.
\begin{table}[!h]
    \centering
    \resizebox{\columnwidth}{!}{%
    \begin{tabular}{cccccc}
    \toprule
    \textbf{Approach} & \textbf{B1} & \textbf{B2} & \textbf{B3} & \textbf{B4} & \textbf{Global} \\
    \midrule
    Original & 3.61 & 3.42 & 2.12 & 1.47 & 38.48\\ 
    \midrule
    Layer Folding & \bf 1.01 & 7.52 & 5.19 & 1.01 & 39.81 \\ 
    \midrule
    EGP & 3.83 & 3.70 & \bf 1.01 & \bf 1.01 & 14.17 \\
    \midrule
    NEPENTHE & 4.53 & 3.03 & \bf 1.01 & \bf 1.01 & 13.73 \\
    \midrule
    EASIER & 4.72 & 1.44 & 1.35 & 2.69 & 24.68 \\ 
    \midrule
    LaCoOT ($\lambda=5$) & \bf 1.01 & \bf 1.04 & 1.10 & 1.18 & \bf 1.36 \\
    \bottomrule
    \end{tabular}
}
\caption{Lipschitz constants for ResNet-18 on CIFAR-10.}
\label{tab:Lipschitz}
\end{table}
\section{Comparison with structured pruning}
\label{sec:structured_pruning}

To complete the comparisons carried out in Sec. 5, we compare in this section the effectiveness of LaCoOT with respect to a traditional model pruning method : Depgraph~\cite{fang2023depgraph} in Tab.~\ref{tab:depgraph}. LaCoOT outperforms DepGraph and achieves superior latency reduction. DepGraph's low performance is due to the absence of retraining after pruning (to fairly compare to us).
Furthermore, since ~\cite{liu2021discrimination,diaopruning,fang2023depgraph} perform channel pruning while LaCoOT removes entire layers, these methods operate at different levels of granularity. Rather than addressing the exact same problem, they are complementary: applying DepGraph on top of LaCoOT (as a refinement at finer granularity) yields even greater latency gains.

\begin{table}[!h]
\centering
\resizebox{\columnwidth}{!}{%
    \centering
    \begin{tabular}{c|c|c|c}
    \toprule
        \textbf{Approach} & \textbf{top-1 [\%]} & \textbf{MACs [M]} & \textbf{Latency} \\
    \midrule
        Original & 91.77 & 140.19 & 100\% \\ 
    \midrule
        LaCoOT ($\lambda=5$) & 90.99 & 64.69 & -38\% \\
        Depgraph (0.3) & 59.40 & 70.96 & -11\% \\
    \midrule
        LaCoOT ($\lambda=5$) + Depgraph (0.2) & 90.96 & 57.22 & -45\% \\
        LaCoOT ($\lambda=5$) + Depgraph (0.3) & 89.27 & 49.63 & -49\% \\ 
    \bottomrule
    \end{tabular}
        }
\caption{Depgraph vs. LaCoOT for ResNet-18 on CIFAR-10.}
\label{tab:depgraph}
\end{table}
\section{A closer look at generated samples}

\begin{figure*}[ht]
    \centering
    \begin{subfigure}[b]{\linewidth}
    \centering
        \includegraphics[width=0.91\linewidth]{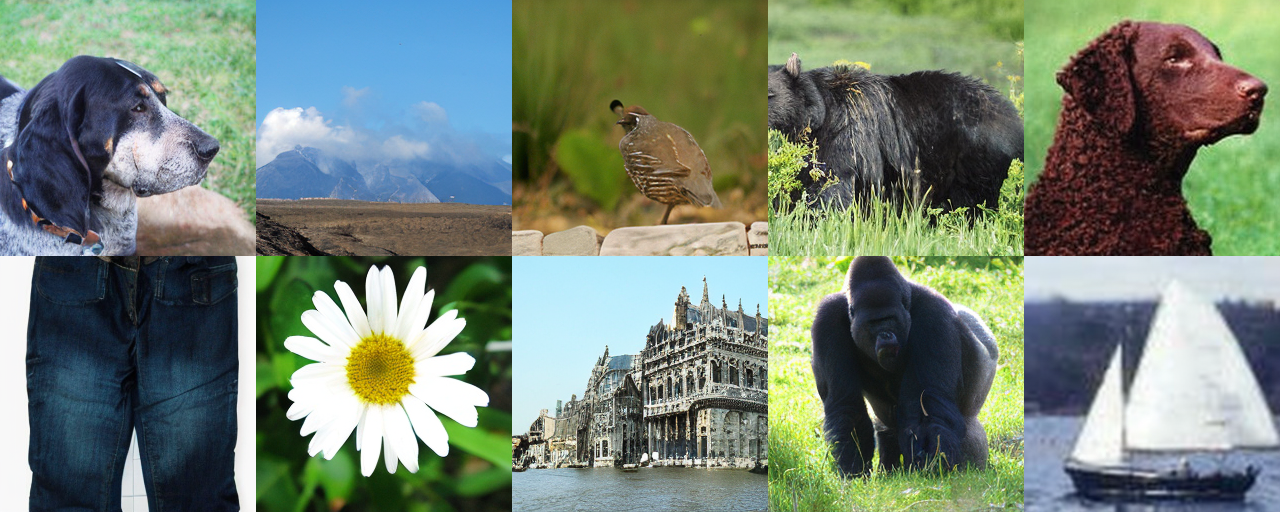}
        \caption{Samples generated from a pre-trained DiT-XL/2.}
        \label{fig:DiT_supp_baseline}
    \end{subfigure}\hfill
    \begin{subfigure}[b]{\linewidth}
    \centering
       \includegraphics[width=0.91\linewidth]{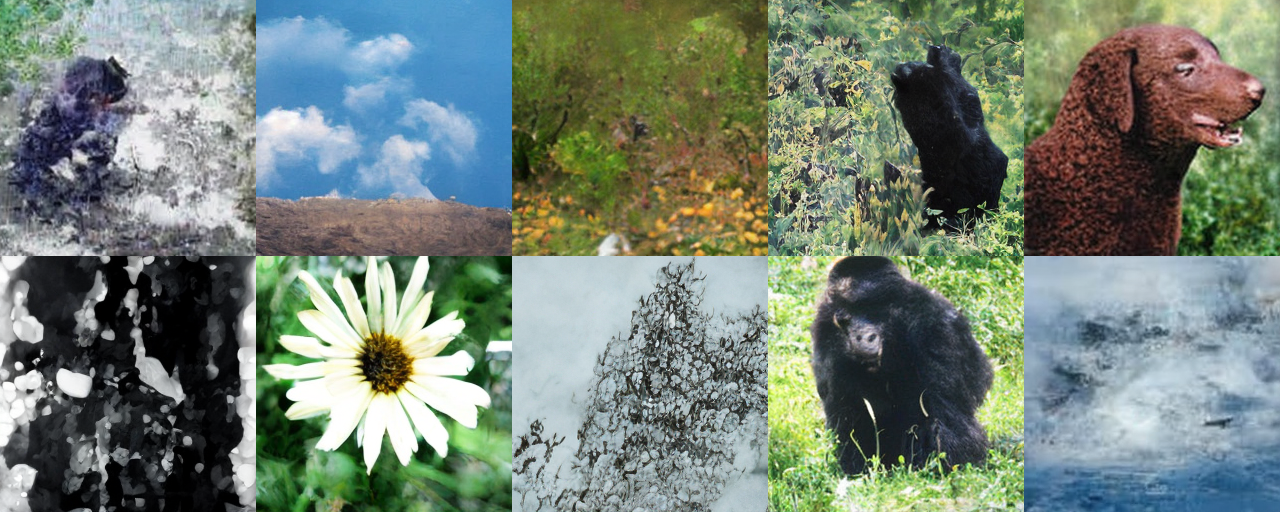}
        \caption{Samples generated from a DiT-XL/2 with two DiT blocks removed, without LaCoOT. The generated content tends to be indiscernible.}
        \label{fig:DiT_supp_without}
    \end{subfigure}\hfill
    \begin{subfigure}[b]{\linewidth}
    \centering
        \includegraphics[width=0.91\linewidth]{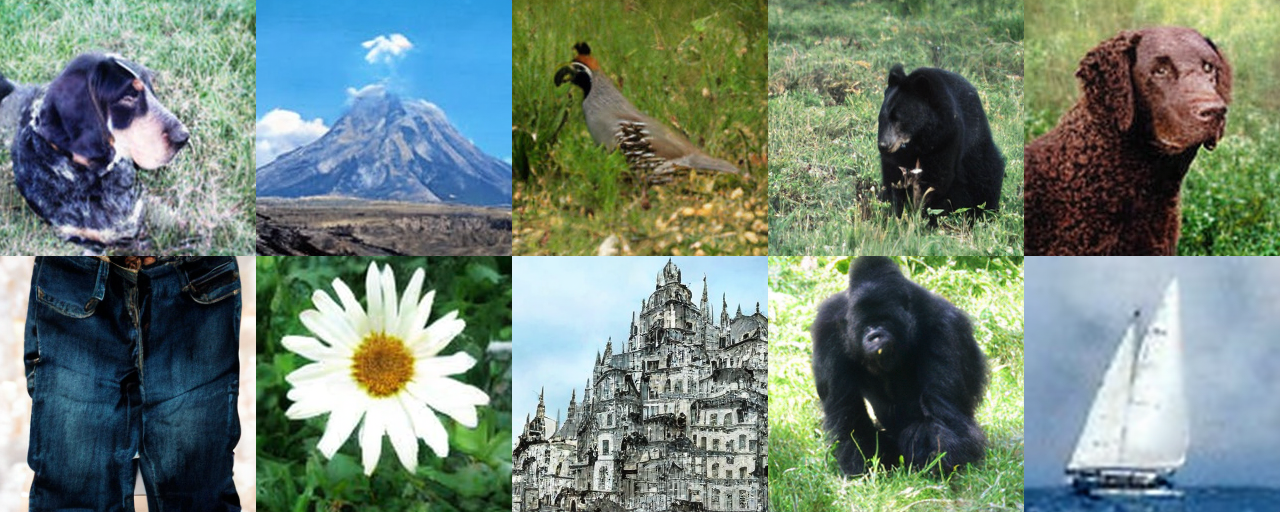}
        \caption{Samples generated from a DiT-XL/2 finetuned with LaCoOT ($\lambda=1e\!\mathpunct{-}\!4$) with two DiT blocks removed.}
        \label{fig:DiT_supp_lacoot}
    \end{subfigure}
    \caption{Generated samples from different configurations of a DiT-XL/2. When finetuned with LaCoOT, when two DiT blocks are removed, the generated content is better preserved. Indeed, the removal of blocks is completely destroying  generated images in absence of the regularization, while the generated content is better preserved with its use.}
    \label{fig:DiT_supp_main}
\end{figure*}

We display in Fig.~\ref{fig:DiT_supp_main} some generated samples from the pre-trained DiT-XL/2 (Fig.~\ref{fig:DiT_supp_baseline}), a DiT-XL/2 with two DiT blocks removed without the use of LaCoOT (Fig.~\ref{fig:DiT_supp_without}), and from a DiT-XL/2 finetuned with LaCoOT($\lambda=1e\!\mathpunct{-}\!4$) with two DiT blocks removed (Fig.~\ref{fig:DiT_supp_lacoot}).

While the removal of blocks is completely destroying  generated images in absence of the regularization, the generated content is better preserved when the DiT-XL/2 is fine-tuned with our method on 5k training steps. 
Indeed, the resulting images are much better than the the generated images produced without LaCoOT.
However, we can observe a little loss in visual fidelity with respect to the pre-trained DiT-XL/2. For instance, although we can still perceive the buildings in the third image of the second column in Fig.~\ref{fig:DiT_supp_lacoot}, we can no longer discern the lake in the foreground compared to the pre-trained model image in Fig.~\ref{fig:DiT_supp_baseline}.
Nevertheless as it required only a few finetuning steps and given the quality of the generated samples with our method compared to without, we believe that our approach LaCoOT can be applied and suitable for foundation models. 
\section{Ablation Study}

In this section, we conduct multiple ablation studies. First, we explore in Sec.~\ref{subsec:ablation_1} the impact of using the Max-Sliced Wasserstein Distance, or the the Sliced Wasserstein Distance as a regularization in our method.
Second, we evaluate the impact of the number of projections in Sec.~\ref{subsec:ablation_2} and the batch size in Sec.~\ref{subsec:ablation_3} toward LaCoOT success.
Finally, in Sec.~\ref{subsec:ablation_4}, we compare our method LaCoOT replacing the Max-Sliced Wasserstein Distance with other existing metrics to quantify differences between distributions.

\subsection{Theoretical guarantees versus practical benefits}
\label{subsec:ablation_1}
From the POT library~\cite{pot}, two sliced OT distances can be used in our proposed regularization strategy. We propose here to explore the impact of using the Max-Sliced Wasserstein Distance (MSWD), or the the Sliced Wasserstein Distance (SWD) as a regularization in our method.

From a theoretical perspective, it is preferable to use the MSWD as it guarantees convergence~\cite{Deshpande2019}. Indeed, the MSWD minimizes the worst-case difference in distribution between the two measures over all possible projections. Since the MSWD is a global measure over all slices, it enforces convergence in the full measure space. This makes it a robust and convergent method for comparing distributions, ensuring that all possible distances are minimized when the maximum distance is minimized.

\begin{figure}[!h]
    \centering
    \includegraphics[width=\columnwidth]{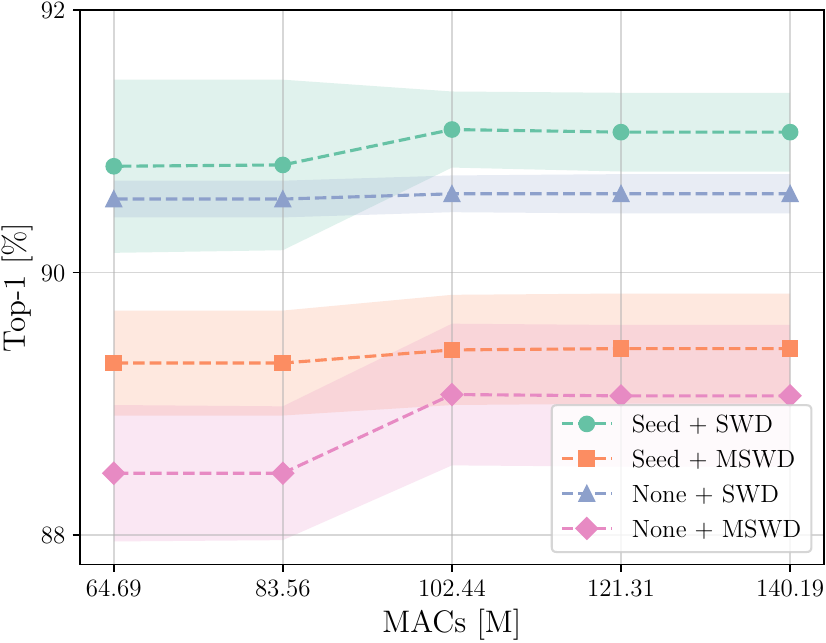}
    \caption{MSWD vs. SWD and impact of seeding the generator for the projections for a ResNet-18 trained on CIFAR-10 with LaCoOT ($\lambda=5)$. SWD yields better results than MSWD.}
    \label{fig:Ablation_seed_maxSWD}
\end{figure}

Moreover, when performing projections to calculate the Wasserstein Distance, we evaluate the impact of seeding the generator. Specifically, we investigate whether initializing the random seed for the generator during the projection process affects the stability or performance of our model.

This leads to four distinct configurations:
\begin{itemize}
    \item ``Seed + SWD'' refers to the case where the SWD is used as a regularizer in our framework, and the generator is seeded during projections; 
    \item ``Seed + MSWD'' refers to the case where the MSWD is used as a regularizer in our framework, and the generator is seeded during projections;
    \item ``None + SWD'' refers to the case where the SWD is used as a regularizer in our framework, and the generator is not seeded during projections, allowing for randomness to influence the projection directions;
    \item ``None + MSWD'' refers to the case where the MSWD is used as a regularizer in our framework, and the generator is not seeded during projections, allowing for randomness to influence the projection directions.
\end{itemize}

Since the unseeded approach may expose the model to more generalization across different slices, and that MSWD provides theoretical convergence guarantees, we use the ``None + MSWD'' configuration in all our experiments except where otherwise stated.

Fig.~\ref{fig:Ablation_seed_maxSWD} displays the results of the ablation over the 4 configurations on a ResNet-18 trained on CIFAR-10 with our method LaCoOT with $\lambda=5$. Since variability between runs can occur, we report standard deviations over 5 runs.

Interestingly, despite offering theoretical convergence guarantees, the use of MSWD as a regularizer yields worse results compared to the use of the SWD. 
Moreover, looking at the standard deviations, we can observe that the ``None + SWD'' configuration display the lowest, which shows its stability. 
Thus, we draw the reader's attention to the fact that better results can be obtained in practice if one allow himself to dispense with the theoretical convergence guarantees. 

\subsection{Ablation on the number of projections}
\label{subsec:ablation_2}
\begin{figure}[!h]
    \centering
    \includegraphics[width=\columnwidth]{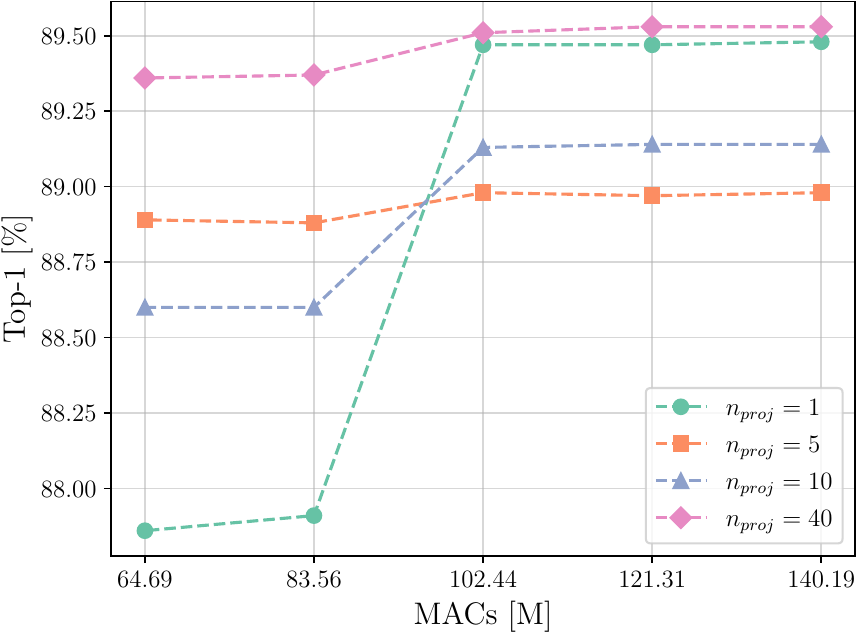}
    \caption{Ablation on the number of projection $n_{proj}$ for a ResNet-18 trained on CIFAR-10 with LaCoOT ($\lambda=5)$.}
    \label{fig:Ablation_n_proj}
\end{figure}

In this subsection, we evaluate the impact of the number of projections $n_{proj}$ toward LaCoOT success.
Indeed, Fig.~\ref{fig:Ablation_n_proj} shows the results achieved for LaCoOT($\lambda=5)$ when lowering the number of projections used to calculate the MSWD.

While for the extreme case ($n_{proj}=1$), a drop in performance is observed when blocks are removed and MACs reduced, we can already obtain decent results with $n_{proj}=5$. 
Indeed, it appears that the number of projections plays a role in the trade-off between the model's original performance (at 140.19 MACs) and the possibility of removing layers without performance loss. In fact, $n_{proj}=40$ showcases the best results with the best performance at lower MACs, and a very slight loss of performance with respect to its original counterpart at 140,19 MACs. 

\subsection{Ablation on the batch size}
\label{subsec:ablation_3}
\begin{figure}[!h]
    \centering
    \includegraphics[width=\columnwidth]{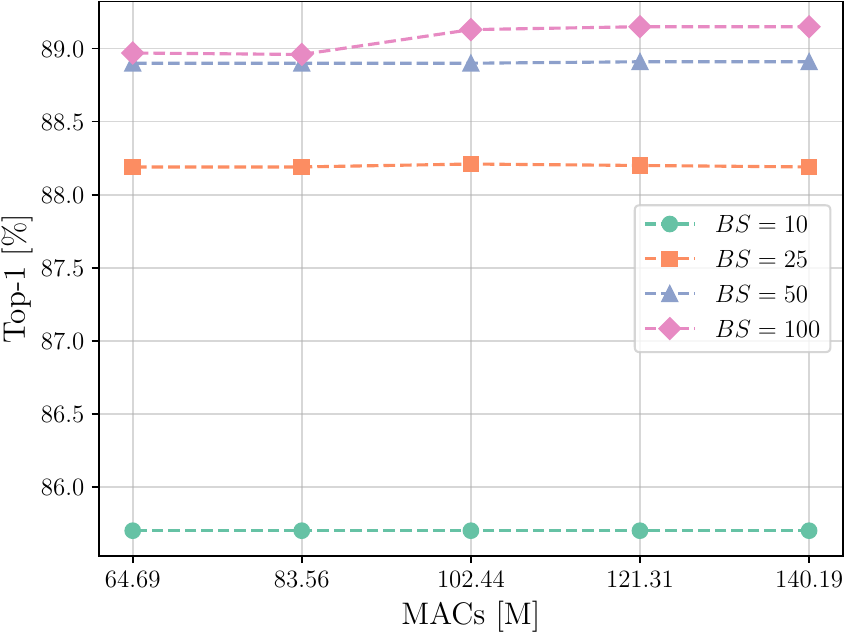}
    \caption{Ablation on the batch size $BS$ for a ResNet-18 trained on CIFAR-10 with LaCoOT ($\lambda=5)$.}
    \label{fig:Ablation_batch_size}
\end{figure}

In this subsection, we evaluate the impact of the batch size $BS$ on LaCoOT results.
Indeed, Fig.~\ref{fig:Ablation_batch_size} shows the results achieved for LaCoOT($\lambda=5)$ when lowering the batch size used to calculate the MSWD.

Looking at the results, it appears evident that reducing the batch size produces worse results. Although performance remains constant (or presents a slight decrease for $BS=100$) when layers are removed and MACs are reduced, it appears evident that the smaller the batch size, the lower the performance of the original model (at 140.19 MACs).
Hence, whenever possible, LaCoOT should always be applied with a sufficiently large batch size that can fit in the memory of the used computing resources.

\subsection{Comparison with other metrics}
\label{subsec:ablation_4}
\begin{figure}[!h]
    \centering
    \includegraphics[width=\columnwidth]{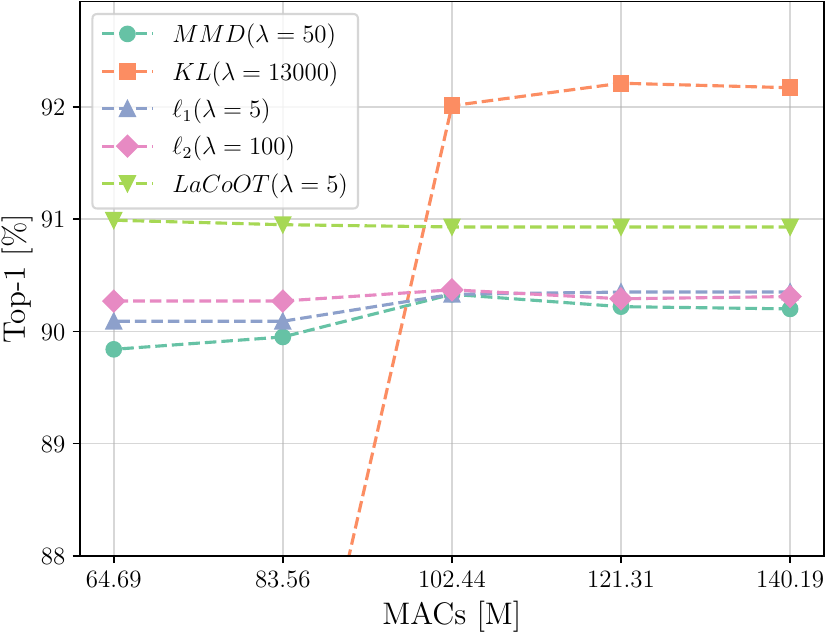}
    \caption{Ablation on LaCoOT replacing the proposed regularization with $\ell_1$, $\ell_2$, MMD or KL divergence, for a ResNet-18 trained on CIFAR-10.}
    \label{fig:Ablation_other_method}
\end{figure}

In this subsection, we compare our method LaCoOT using other existing metrics to quantify differences between distributions. 
Indeed, the MSWD regularization can be replaced by the $\ell_1$ distance, the $\ell_2$ distance, the Maximum Mean Discrepancy (MMD) or the Kullback-Leibler (KL) Divergence.
For each comparison, we show the best configuration of $\lambda$ yielding the best results for the trade-off between top-1 performance and MACs. Indeed, a grid search on $\lambda$ is carried out to find the best trade-offs. 
Fig.~\ref{fig:Ablation_other_method} displays the results for a ResNet-18 trained on CIFAR-10.

On the one hand, while for high MACs, the KL divergence shows its competitiveness, the performance drops dramatically as blocks are removed and MACs reduced.
On the other hand, we can observe that $\ell_1$, $\ell_2$ and MMD obtain similar performance/complexity trade-off with relatively few performance drops.
Our method LaCoOT with the MSWD outperforms the other compared metrics for lower MACs. 

While the choice of metric seems to have very little impact on the performance obtained, we draw the reader's attention to the fact that this is due to the idea we are proposing. Indeed, minimizing the distance between feature distributions of successive layers appears to be more important to reduce the depth of DNNs than the choice of the metric itself.
Thus, the other metrics presented here can also be used as regularizations in our method, but these may produce worse results.
\section{Training from scratch with lower initial depth}

While having an oracle baseline on each setup is computationally expensive as brute-force research for all the combinations (+full retraining) is required, we present in Tab.~\ref{tab:NAS_like} below the performance of 16 residual networks trained from scratch on CIFAR-10 following the set of Tab.~\ref{tab:learning_strategies}.
Indeed, we remove at initialization (a combination of) layers inside a ResNet-18 and train the network from scratch. We call this  the ``brute-force approach''.

\begin{table}[!h]
    \centering
    \resizebox{\columnwidth}{!}{%
    \begin{tabular}{cccc}
    \toprule
        \textbf{Combination} & \textbf{\# B. Rem.} & \textbf{Top-1 [\%]} & \textbf{MACs [M]} \\ 
    \midrule
        2222 (Original) & 0 & 91.77 & 140.19 \\
    \midrule
        2221 & 1 & 91.86 & 121.31 \\ 
        2212 & 1 & 91.29 & 121.31 \\ 
        1222 & 1 & 90.82 & 121.31 \\ 
        2122 & 1 & 90.59 & 121.31 \\ 
        2211 & 2 & 91.96 & 102.44 \\ 
        1221 & 2 & 91.58 & 102.44 \\ 
        2112 & 2 & 91.45 & 102.44 \\ 
        2121 & 2 & 91.27 & 102.44 \\ 
        1212 & 2 & 91.25 & 102.44 \\ 
        1122 & 2 & 91.02 & 102.44 \\ 
        2111 & 3 & 91.78 & 83.56 \\ 
        1211 & 3 & 91.64 & 83.56 \\ 
        1121 & 3 & 90.94 & 83.56 \\ 
        1112 & 3 & 90.88 & 83.56 \\ 
        1111 & 4 & 91.45 & 64.69 \\
    \midrule
        LaCoOT($\lambda=5$) & 4 & 90.99 & 64.69 \\ 
        LaCoOT($\lambda=5$) + retraining & 4 & 91.42 & 64.69 \\
    \bottomrule
    \end{tabular}
    }
    \caption{ResNet-18 trained from scratch on CIFAR-10 with layers removed initially. For a given combination, we associate the number of blocks removed (\# B. Rem.), the top-1 performance and associated MACs at inference.}
    \label{tab:NAS_like}
\end{table}

Our approach LaCoOT is achieving comparable performance compared to these models. However, while the  ``brute-force approach'' has to perform 16 separate trainings, LaCoOT can produce the same 16 subnetworks in one training only, hence being very efficient. By further retraining the pruned architecture, we recover performance, comparable to the original model, as shown in the last line.
\section{Details on the learning strategies employed}
\label{appendix:details}

\noindent\textbf{Image Classification.}
The training hyperparameters used in the experiments are presented in Table~\ref{tab:learning_strategies}. Our code is available at \url{https://github.com/VGCQ/LaCoOT}.

\begin{table*}[t]
    \centering
    \resizebox{0.9\textwidth}{!}{%
    \begin{tabular}{c c c c c c c c c c}
        \toprule
        \bf Model & \bf Dataset & \bf Epochs & \bf Batch & \bf Opt. & \bf Mom. & \bf LR & \bf Milestones & \bf Drop Factor & \bf Weight Decay \\
        \midrule
         ResNet-18 & CIFAR-10 & 160 & 128 & SGD & 0.9 & 0.1 & [80, 120] & 0.1 & 1e-4 \\
         Swin-T & CIFAR-10 & 160 & 128 & SGD & 0.9 & 0.001 & [80, 120] & 0.1 & 1e-4\\
         MobileNetv2 & CIFAR-10 & 160 & 128 & SGD & 0.9 & 0.1 & [80, 120] & 0.1 & 1e-4\\
         \midrule
         ResNet-18 & Tiny-ImageNet-200 & 160 & 128 & SGD & 0.9 & 0.1 & [80, 120] & 0.1 & 1e-4 \\
         Swin-T & Tiny-ImageNet-200 & 160 & 128 & SGD & 0.9 & 0.001 & [80, 120] & 0.1 & 1e-4 \\
         MobileNetv2 & Tiny-ImageNet-200 & 160 & 128 & SGD & 0.9 & 0.1 & [80, 120] & 0.1 & 1e-4 \\
         \midrule
         ResNet-18 & PACS & 30 & 16 & SGD & 0.9 & 0.001 & [24] & 0.1 & 5e-4 \\
         Swin-T & PACS & 30 & 16 & SGD & 0.9 & 0.001 & [24] & 0.1 & 5e-4 \\
         MobileNetv2 & PACS & 30 & 16 & SGD & 0.9 & 0.001 & [24] & 0.1 & 5e-4 \\
         \midrule
         ResNet-18 & VLCS & 30 & 16 & SGD & 0.9 & 0.001 & [24] & 0.1 & 5e-4 \\
         Swin-T & VLCS & 30 & 16 & SGD & 0.9 & 0.001 & [24] & 0.1 & 5e-4 \\
         MobileNetv2 & VLCS & 30 & 16 & SGD & 0.9 & 0.001 & [24] & 0.1 & 5e-4 \\
         \midrule
         ResNet-18 & Flowers-102 & 50 & 16 & Adam &  & 1e-4 & ~ & ~ & 0  \\
         Swin-T & Flowers-102 & 50 & 16 & Adam &  & 1e-4 & ~ & ~ & 0 \\
         MobileNetv2 & Flowers-102 & 50 & 16 & Adam &  & 1e-4 & ~ & ~ & 0 \\
         \midrule
         ResNet-18 & DTD & 50 & 16 & Adam &  & 1e-4 & ~ & ~ & 0  \\
         Swin-T & DTD & 50 & 16 & Adam &  & 1e-4 & ~ & ~ & 0 \\
         MobileNetv2 & DTD & 50 & 16 & Adam &  & 1e-4 & ~ & ~ & 0 \\
         \midrule
         ResNet-18 & Aircraft & 50 & 16 & Adam &  & 1e-4 & ~ & ~ & 0  \\
         Swin-T & Aircraft & 50 & 16 & Adam &  & 1e-4 & ~ & ~ & 0 \\
         MobileNetv2 & Aircraft & 50 & 16 & Adam &  & 1e-4 & ~ & ~ & 0 \\
          \bottomrule
    \end{tabular}
    }
    \caption{The different employed learning strategies.}
    \label{tab:learning_strategies}
\end{table*}

CIFAR-10 is augmented with per-channel normalization, random horizontal flipping, and random shifting by up to four pixels in any direction.
For the datasets of DomainBed, the images are augmented with per-channel normalization, random horizontal flipping, random cropping, and resizing to 224. The brightness, contrast, saturation, and hue are also randomly affected with a factor fixed to 0.4.
Tiny-ImageNet-200 is augmented with per-channel normalization and random horizontal flipping.
Moreover, the images of Flowers-102 are augmented with per-channel normalization, random horizontal and vertical flipping combined with a random rotation, and cropped to 224. DTD and Aircraft are augmented with random horizontal and vertical flipping, and with per-channel normalization.

Following~\cite{liao2023can} and~\cite{quetu2024dsd2}, on CIFAR-10 and Tiny-ImageNet-200, all the models are trained for 160 epochs, optimized with SGD, having momentum 0.9, batch size 128, and weight decay 1e-4. The learning rate is decayed by a factor of 0.1 at milestones 80 and 120. The initial learning rate ranges from 0.1 for ResNet-18 and MobileNetv2, 
to 1e-3 for Swin-T.
Moreover, on PACS and VLCS, all the models are trained for 30 epochs, optimized with SGD, having momentum 0.9, a learning rate of 1e-3 decayed by a factor 0.1 at milestone 24, batch size 16, and weight decay 5e-4.
Furthermore, on Aircraft, DTD, and Flowers-102, all the models are trained following a transfer learning strategy. Indeed, each model is initialized with its pre-trained weights on ImageNet, trained for 50 epochs, optimized with Adam, having a learning rate 1e-4 and batch size 16.

The experiments were mostly performed using an NVIDIA RTX 3090. 

\noindent\textbf{Image Generation.} 
DiT-XL/2 at 256 $\!\times\!$ 256 image resolution is fine-tuned on ImageNet for 5k training steps on 3 NVIDIA L40S using AdamW, no weight decay, with a global batch size of 60 and a learning rate 1e-4. 
Only horizontal flips were used to augment the training set. Following common practice in the generative modeling literature, we maintain an exponential moving average (EMA) of DiT weights over training with a decay of 0,9999. All results reported use the EMA model. The pre-trained model is taken from the original paper~\cite{peebles2023scalable} and the diffusion was done using the same details as in the original paper. 
We evaluate the quality of the generated samples with Fréchet Inception Distance (FID)~\cite{heusel2017gans}, the standard metric for evaluating generative models of images. Following convention, we report FID-50k using 250 sampling steps with clean-fid~\cite{parmar2022aliased} and classifier-free guidance scale of 1,5.

\end{document}